\documentclass{article}

\usepackage{arxiv}

\usepackage[utf8]{inputenc} 
\usepackage[T1]{fontenc}    
\usepackage{hyperref}       
\usepackage{url}            
\usepackage{booktabs}       
\usepackage{amsfonts}       
\usepackage{nicefrac}       
\usepackage{microtype}      
\usepackage{lipsum}		
\usepackage{graphicx}
\usepackage{natbib}
\usepackage{doi}
\usepackage{multicol}
\usepackage{multirow}
\usepackage{caption}
\usepackage{amsmath}
\usepackage{enumitem}

\title{Realistic Counterfactual Explanations by Learned Relations}

\date{} 					

\author{{Xintao Xiang} \\
	School of Computer Science\\
	Australian National University\\
	Canberra, ACT \\
	\texttt{xintao.xiang@anu.edu.au} \\
	\And
	{Artem Lenskiy} \\
    School of Computer Science\\
	Australian National University\\
	Canberra, ACT \\
	\texttt{artem.lenskiy@anu.edu.au} \\
}

\renewcommand{\shorttitle}{Realistic Counterfactual Explanations}

\hypersetup{
pdftitle={Counterfactual Explanations Constrained in Learned Relations},
pdfauthor={Xintao Xiang, Artem Lenskiy},
pdfkeywords={Explainable machine learning,  realistic counterfactual explanations},
}

\begin{document}
\maketitle

\begin{abstract}
Many existing methods of counterfactual explanations ignore the intrinsic relationships between data attributes and thus fail to generate realistic counterfactuals. Moreover, the existing models that account for relationships require domain knowledge, which limits their applicability in complex real-world applications. In this paper, we propose a novel approach to realistic counterfactual explanations that preserve the relationships and minimise experts' interventions. The model directly learns the relationships by a variational auto-encoder with minimal domain knowledge and then learns to perturb the latent space accordingly. We conduct extensive experiments on both synthetic and real-world datasets. The experimental results demonstrate that the proposed model learns relationships from the data and preserves these relationships in generated counterfactuals. In particular, it outperforms other methods in terms of Mahalanobis distance, and the constraint feasibility score.
\end{abstract}

\keywords{Explainable machine learning,  realistic counterfactual explanations}

\section{Introduction}
In light of widespread of deep learning models in numerous engineering and medical applications, being able to explain these models has become an important issue\cite{trumbelj2014explaining, datta2016algorithmic}. There are two directions that focus on providing explanations to the models and consequent decisions made by them. The first group focuses on explanations by means of local classification boundaries (e.g. LIME \cite{ribeiro2016why}) to explain why a give sample was classified this way, the other one focuses on an entire model and present importance of each feature (e.g. SHAP \cite{lundberg2017a}). Both approaches assign each predictor an importance weight to a particular prediction. For example, if a bank loan was rejected, then the bank can employ the SHAP values to explain the reasons behind this decision i.e. to let the client know that his/her salary was an important factor in declining the loan. 
Although these approaches are useful in helping to uncover mysterious machine learning models that are often treated as black boxes, there are a number of applications where information about feature importance is not sufficient and counterfactual explanations are more favorable. Going back to the example of credit predication: a client whose loan application is rejected, would like to know how can he/she improve the chances in getting the application approved next time, rather than just being presented with the importance of each factor. In other words,  rather than just explaining what factors played a role in rejecting the loan, provide a recommendation on what exactly needs to be done to reach a positive decision next time. The method of counterfactual explanations focuses on answering the question  - \textit{How could I get the desired target, given that I am in the current situation with respect to the employed decision model?}

In a recent review on methods for counterfactual explanations \cite{verma2020counterfactual}, the nearest counterfactual explanation was identified as one of the most popular methods in generating counterfactuals. The method provides a nearest explanation with a desired output \cite{wachter2018counterfactual} by optimising the following criterion
\begin{equation}
    \label{e1}
	\mathop{\arg \text{min}}\limits_{\mathbf{x}^{CF}}\
	\lambda(f(\mathbf{x}^{CF})-y^{CF})^2+d(\mathbf{x},\mathbf{x}^{CF}),
\end{equation}
where $\mathbf{x}$ is a sample, $\mathbf{x}^{CF}$ is a corresponding generated counterfactual, $y^{CF}$ is the target label for counterfactual and $\lambda$ is a scale factor that controls a trade-off of accuracy and closeness. Some other constraints such as sparsity and feasibility can be added to generate more favorable explanations \cite{dandl2020multi, karimi2020model}. However, recent research raises concerns about feasibility in counterfactual explanations \cite{mahajan2019preserving} in real-world applications. For example, in a medical application where we are giving suggestions to patients, we should consider that the blood pressure and blood mass index (BMI) are positively correlated. Therefore, we should not suggest decreasing blood pressure while increasing BMI. Indeed, some methods \cite{karimi2020algorithmic, karimi2020algorithmic2, downs2020cruds} have been proposed to address this very problem, however there are some downsides.

\noindent\textbf{Existing models and their drawbacks} \textit{Existing methods all assume complete or at least partial domain knowledge, which is not always can be put into practice as part of real-world applications}. Mahajan et al. proposed to modify the optimisation criterion in such a way that expert knowledge (i.e. prior examples or external oracle) is taken into account and the model is retrained accordingly\cite{mahajan2019preserving}.  Downs et al. proposed to use conditional subspace variational auto-encoder that generates multiple recourse satisfying underlying structure of the data with end-user specified constraints \cite{downs2020cruds}. Karimi et al. \cite{karimi2020algorithmic, karimi2020algorithmic2} assumed that at least partial causal graph is available during training. While these approaches indeed improves the quality of generated counterfactuals, these methods requires domain knowledge which is often hard to obtain, especially in areas such as biology or medicine that frequently deal with high-dimensional feature spaces. Practically,  \textit{only labeled data is available} without or with minimal expert input. This practical limitation motivates us to search for methods of causal inference \cite{pearl2016causal} in order to infer the inner relationships between observed variables.

In this paper, we propose a counterfactual explanation method that assumes no prior knowledge on relationships between the variables in an application domain. To learn the relationships between data attributes, our method first trains a variational auto-encoder (VAE) that is based on DAG-GNN \cite{yu2019dag}. Then, the proposed method learns a modulation network that takes the query data points and corresponding target labels as inputs, and perturb the latent representations of the data points so that counterfactuals can be constructed by the perturbed latent representations. To make sure that the generated counterfactuals preserve the learned relationships, we use adversarial training to match the counterfactual latent distribution to the query data latent distribution. The proposed approach thus can provide a realistic and satisfactory recommendation by generating counterfactuals without any prior domain knowledge.

The main contributions of this work are as follows:
\begin{itemize}[leftmargin=*]
    \item We propose a novel model that considers relationships between attributes and gives realistic counterfactual explanations of machine learning model.
    \item The model does not need any prior domain knowledge on predictors' correlation and is thus more flexible and useful in complex real-world settings where prior knowledge is difficult to obtain.
    \item The extensive experiments on both synthetic and real-world datasets demonstrate the feasibility of the proposed model.
\end{itemize}

\section{Background and Related Work}
\noindent\textbf{Causal relationships.} Causal relationships are commonly represented by directed acyclic graphs (DAG): $\mathcal{G}=(\mathcal{V},\mathcal{E})$, with $\mathcal{V}$ being a set of nodes representing attributes $\{v_1,...,v_L\}$, and $\mathcal{E}$ denoting the directed edges corresponding to causal relations. The causal relations between variables can be modelled by a structural causal model (SCM) \cite{elements_causal}:
\begin{equation}
	\label{SCM}
	v_j = f_j(\mathcal{P}_j,\epsilon_j) \ \ \ \ j=1,...,L,
\end{equation}
where $\mathcal{P}_j$ denotes parents of $v_j$ (the nodes that point to node $v_j$) and $\epsilon_j$ is its corresponding exogenous variable. The problem has been extensively studied over the past decades \cite{spirtes2000causation,mooij2016distinguishing,glymour2019review,zhu2020causal}. In this paper, our proposed method builds on an extension of VAE and implements DAG-GNN in accordance to \cite{yu2019dag}. The method learns a DAG from data which represents the relationships between data attributes. Note that here we consider undirected relationships, as in real-world applications it is impossible to intervene a variable independently of its causes and hence to infer a direct relationship. \cite{mahajan2019preserving}.

\noindent\textbf{Variational auto-encoder.} VAE proposed by \cite{kingma2014auto} is a Bayesian neural network that learns to represent an observation $\mathbf{x}$ as a latent variable $\mathbf{z}$. The model aims at maximizing the likelihood $\texttt{log}\ p(\mathbf{x})$. While the actual posterior $p(\mathbf{x}|\mathbf{z})$ is intractable, VAE uses variational posterior $q(\mathbf{z}|\mathbf{x})$ to approximate it. The training objective can be constructed by the Evidence Lower Bound (ELBO):
\begin{equation}
    \mathcal{L}(\theta,\phi;\mathbf{x},\mathbf{z}) = \mathbb{E}_{q_{\phi}(\mathbf{x}|\mathbf{z})}\big[\log p_{\theta}(\mathbf{x}|\mathbf{z})\big] - \texttt{D}_{KL}\big(q_{\phi}(\mathbf{z}|\mathbf{x})||p(\mathbf{z})\big)
\end{equation}
where $\theta$ and $\phi$ are neural network parameters. In this paper, we build a generative model based on VAE, specifically built on DAG-GNN \cite{yu2019dag} then generate counterfactuals through the decoder by disturbing the latent representations.

\begin{figure*}[t!]
    \centering
	\includegraphics[width=0.75\textwidth, trim={5.0cm 5.1cm 6cm 3.9cm},clip]{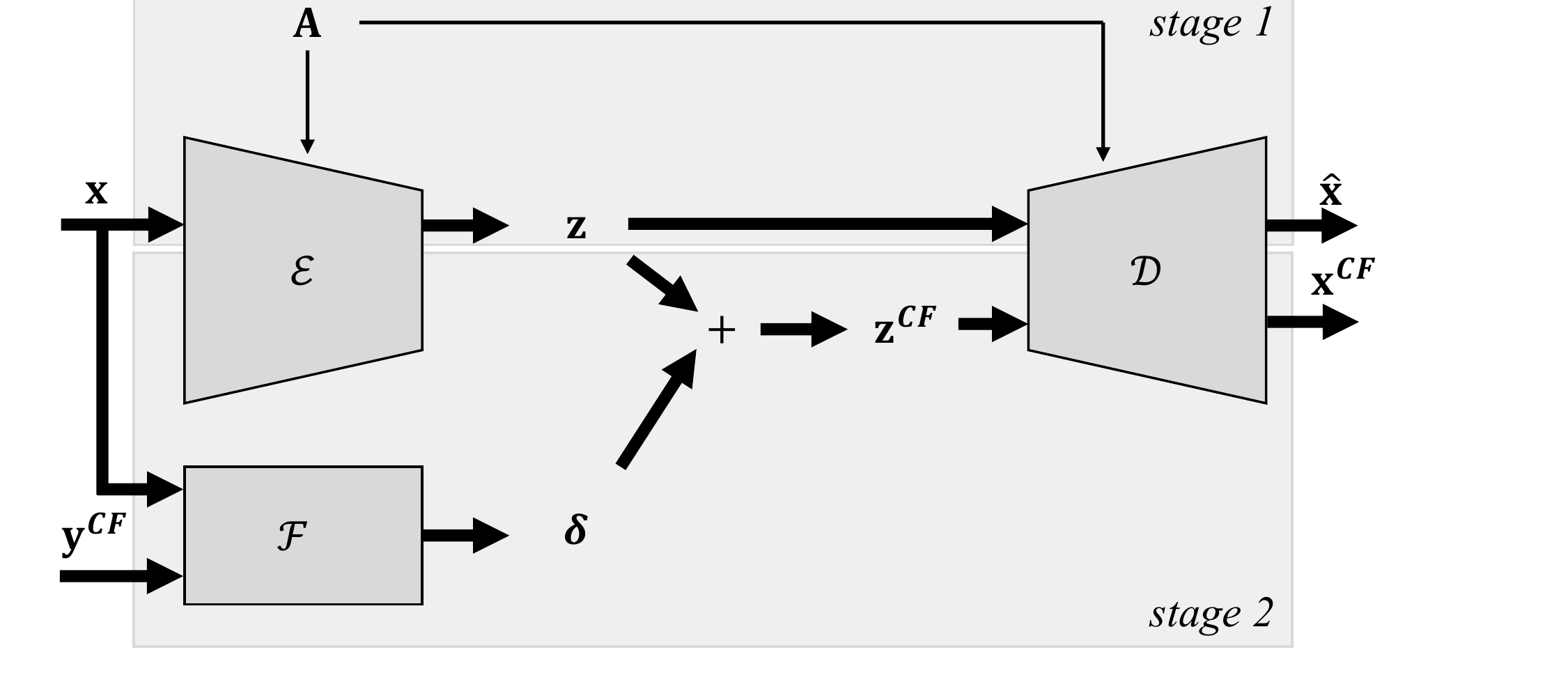}
	\vspace{-2pt}
	\caption{A flowchart model. Stage 1: the encoder and decoder are trained to reconstruct samples. Stage 2: a modulation network is trained to generate disturbance on the latent representation, that is used to construct counterfactuals by the decoder.}
	\vspace{-2pt}
	\label{framework}
\end{figure*}

\noindent\textbf{Counterfactual explanation in machine learning.} Counterfactual explanation in machine learning has received increasing attention in research, especially for high-impact areas such as financial or healthcare \cite{verma2020counterfactual}. A number of methods have been proposed to give an optimal solution to the problem of generating counterfactuals \cite{albini2020relation, kommiya2021towards, parmentier2021optimal}. \cite{wachter2018counterfactual} proposes to generate counterfactuals to explain models by solving the optimization problem formulated by Equation \ref{e1}. Several extensions \cite{dandl2020multi, mothilal2020explaining, grath2018interpretable} have been proposed based on this criterion. Concerns about feasibility, actionability and sparsity in counterfactual explanations have been raised \cite{mahajan2019preserving, karimi2020algorithmic, keane2020good}. A counterfactual explanation is said to be feasible when it captures the relationships between attributes. Actionability aims at providing practical counterfactual and depends on the affordability of individuals on implementing the counterfactual \cite{ferrario2020series}. For example, increasing the age for 20 years is expensive and meaningless for a person in real world. Sparsity applies when shorter explanations are more comprehensible to humans \cite{verma2020counterfactual}, where ideally, the counterfactual changes only a small number of attributes. In this paper, we focus on generating feasible counterfactual. A similar task was investigated by \cite{albini2020relation}, that generated counterfactual explanations for a range of Bayesian network classifiers. While the method does focus on explanations built from relations of influence between variables, it is specifically designed for Bayesian network. On contrary, our method, in theory, can be used for any classifier that allows gradient back propagation. Another method proposed by \cite{mahajan2019preserving} requires expert knowledge. Instead our method can automatically learn the intrinsic relationships between attributes and unlike \cite{karimi2020algorithmic, karimi2020algorithmic2, downs2020cruds}, the proposed method does not rely on prior knowledge.


\section{The model description}
\label{sec:method}

The major issue of generating feasible counterfactuals consists in determining the intrinsic relationships in the data. As stated earlier, existing methods \cite{mahajan2019preserving, karimi2020algorithmic, karimi2020algorithmic2, downs2020cruds} have assumed full or at least some knowledge about the relationships, evidently this assumption is not always practical. Potentially, methods of causal inference, such as \cite{mooij2016distinguishing, zhu2020causal} could be used to determine the causal relationships first and then generate valid counterfactuals by subsequent models of counterfactuals generation. However, such approach requires several models which might result in additional computational costs, especially when methods of exact Bayesian structure discovery are used to learn the structure\cite{Koivisto2005}. The latter are known to be computationally expensive. Therefore, an end-to-end model that directly learns the relationships and generates counterfactuals is preferred.

\begin{figure}[t]
\centering
\includegraphics[width=0.35\textwidth, trim={11cm 6.7cm 12cm 4.5cm}, clip]{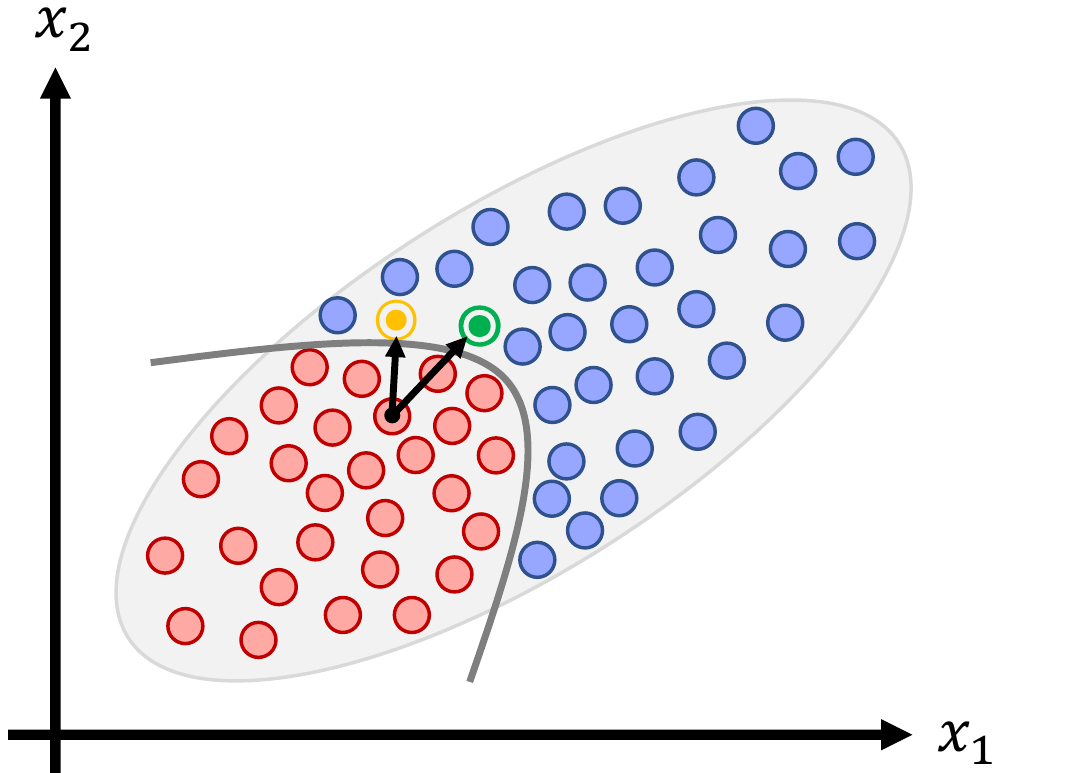}
\vspace{-1pt}
\caption{The red and blue circles are data instances with two attributes positively related. The yellow and green circles are generated counterfactual.}
\vspace{-2pt}
\label{fig:expl}
\end{figure}

Figure \ref{framework} presents an flowchart of the proposed end-to-end model. To learn the relationships, we employ DAG-GNN \cite{yu2019dag} which is designed to learn the directed relationships between the data attributes. This model captures intrinsic relationships between the attributes i.e. the DAG with attributes as nodes. The adjacency matrix helps the model preserve the relationships and the VAE architecture makes it possible to generate valid counterfactuals. One may argue that conditional VAE \cite{sohn2015learning} can also provide in-distribution counterfactuals. However, the in-distribution counterfactuals do not directly imply valid relationships between the attributes. Figure \ref{fig:expl} demonstrates an example, where the yellow counterfactual is in-distribution but not a favorable counterfactual. The DAG-GNN is firstly trained to learn the relationships between the attributes and to reconstruct the samples correctly. Then the counterfactuals are generated by perturbing the learned latent space. To match the perturbed latent distribution to the true latent distribution, we borrow the adversarial idea from adversarial auto-encoder (AAE) \cite{makhzani2015adversarial} and use adversarial training. In the following sections, a rough introduction of DAG-GNN \cite{yu2019dag} is given. Then, the details on how the samples are perturbed in the latent space are provided.

\subsection{Stage 1: Base VAE Pretrain}

VAE serves as the core model for generating counterfactuals. We first pretrain VAE that reconstructs a given sample accurately. We follow the construction process of DAG-GNN \cite{yu2019dag}. In this section, we provide only a brief descriptions of DAG-GNN, for more details, please refer to \cite{yu2019dag}.

Typically, a sample $\mathbf{x}$ is a vector where every attribute is a scalar. The representation can be generalized to $m$ dimensions per attribute. Denote $\mathbf{V} \in \mathbb{R}^{L \times m}$ to represent a sample with $L$ attributes and each attribute has $m$ dimensions. DAG-GNN considers the relationships between attributes by a structural equation model (SEM) of the following  form:
\begin{equation}
	\begin{aligned}
	\mathbf{V} = \mathbf{A}^T\mathbf{V} + \mathbf{Z},
	\end{aligned}
\end{equation}
where $\mathbf{A} \in \mathbb{R}^{L \times L}$ denotes the adjacency matrix of directed relationships between the nodes and $\mathbf{Z} \in \mathbb{R}^{L \times m}$ denotes the exogenous variables. Then, the equation can be represented as $\mathbf{V} = (\mathbf{I} - \mathbf{A}^T)^{-1}\mathbf{Z}$. The decoder's architecture can then be designed in a more general form by taking non-linear transformations:
\begin{equation}
	\mathbf{V} = f^{caus}_2\big((\mathbf{I}-\mathbf{A}^T)^{-1}f^{caus}_1(\mathbf{Z})\big),
\end{equation}
where $f^{caus}_1$ and $f^{caus}_2$ are two parameterized functions that we choose to represent by a multilayer perceptron (MLP). Similarly, the encoder is constructed as:
\begin{equation}
	\label{encoder}
	\mathbf{Z} =f^{caus}_4\big((\mathbf{I}-\mathbf{A}^T)f^{caus}_3(\mathbf{V})\big), 
\end{equation}
where $f^{caus}_3$ and $f^{caus}_4$ can be considered as inverses of $f^{caus}_2$ and $f^{caus}_1$. Note that as we perform non-linear transformations and $\mathbf{Z}$ is treated as latent variables, the choice of hidden dimension $d$ of $\mathbf{Z}$ is arbitrary and so $\mathbf{Z} \in \mathbb{R}^{L \times d}$. Following the VAE construction procedure, the evidence lower bound (ELBO) is given by:
\begin{equation}
	\mathcal{L}_{ELBO}=-\texttt{KL}(q(\mathbf{Z}|\mathbf{V})||p(\mathbf{Z})) + E_{q(\mathbf{Z}|\mathbf{V})}\big[\log p(\mathbf{V}|\mathbf{Z})\big]
\end{equation}

Specifically, the method takes $f^{caus}_1$ as identity mapping, $f^{caus}_2$ as an MLP, and $f^{caus}_3$, $f^{caus}_4$ as an MLP and identity mapping, respectively. For simplicity, the prior is assumed to be the standard normal distribution. In addition to the ELBO loss, an acyclity constraint is added to make $\mathbf{A}$ acyclic:
\begin{equation}
	\texttt{tr}\big[(\mathbf{I}+\alpha \mathbf{A}\circ \mathbf{A})^L\big] - L = 0
\end{equation}
As a result of training by the combined loss listed above, we obtain a VAE that is capable of reconstructing the given samples as well as detecting the relationships between the data attributes.

\subsection{Stage 2: Latent Space Disturbance}

To avoid notation cluttering, in this subsection, we assume all attributes are scalars and denote $\mathbf{z}\in \mathbb{R} ^{d}$ as the latent representation of a sample $\mathbf{x}\in \mathbb{R}^{L}$. Our objective here is to train an effective modulation network denoted by $f^{mod}(\cdot)$ that can perturb the latent representation in a desirable way. Both, a sample $\mathbf{x}$ and a target label $y^{CF}$ are passed to the model. The modulation function $f^{mod}$ that computes the latent code perturbation $\boldsymbol{\delta}$, takes a concatenated vector of $\mathbf{x}$ and $y^{CF}$:
\begin{equation}
	\boldsymbol{\delta} = f^{mod}(\mathbf{x}||y^{CF}),
\end{equation}
where $||$ denotes concatenation. Given a sample $\mathbf{x}$, its latent representation $\mathbf{z}$ is obtained by eq. \ref{encoder}, and the perturbation is added to its latent representation to obtain the latent representation of a corresponding counterfactual $\mathbf{z}^{CF}=\mathbf{z} + \boldsymbol{\delta}$, then the corresponding counterfactual is produced by decoding $\mathbf{z}^{CF}$. Formally, the procedure of generating counterfactual $\mathbf{x}^{CF}$ is given by:
\begin{equation}
	\mathbf{x}^{CF} = f^{caus}_2\big((\mathbf{I}-\mathbf{A}^T)^{-1}f^{caus}_1(\mathbf{z}^{CF})\big).
\end{equation}
Empirically, we observed that a large disturbance $\boldsymbol{\delta}$ at the very beginning of the training made the training unstable. Therefore, to mitigate this effect, the disturbance vector $\boldsymbol{\delta}$ is scaled by a factor $\gamma$ (set to $0.05$).

\noindent\textbf{Objectives functions.} Recall that we impose the requirement for the counterfactual to be valid i.e. to have a predefined target label. Following \cite{mahajan2019preserving}, we introduce a classification loss over the generated counterfactuals as follows:
\begin{equation}
	\mathcal{L}_{class} = \sum^N_{i=1} \text{HingeLoss}(h(\mathbf{x}_i^{CF}), y_i^{CF}, \beta),
\end{equation}
where $h(\cdot)$ is black box model, and the hinge loss is defined over the probability $s_y$ as 
\begin{equation}
	\text{HingeLoss}(h(\mathbf{x}^{CF}), y^{CF}, \beta) = \texttt{max}\{\texttt{max}_{y \neq y^{CF}}\{s_y(\mathbf{x}^{CF})\} - s_{y^{CF}}(\mathbf{x}^{CF}), -\beta \},
\end{equation}
with $\beta$ being a pre-defined margin. This loss function ensures the score of our target label to be higher than any other label at least by $\beta$. Our second objective is to let the counterfactual to be as close to the original sample as possible. Furthermore, since we aim at minimizing the perturbation of the latent code that generates a counterfactual, we penalize large $\boldsymbol{\delta}$ in terms of its $L_2$ norm. Then, the nearest loss is defined as:
\begin{equation}
	\mathcal{L}_{near} = \sum^{N}_{i=1} \big(\texttt{dist}(\mathbf{x}_i, \mathbf{x}_i^{CF}) + ||\boldsymbol{\delta}_i||^2_2\big),
\end{equation}
where $\texttt{dist}(\cdot)$ is a distance function which could be defined by $L_1$ or $L_2$ distances. Since the trained decoder can only decode from a specific distribution (i.e. the latent distribution of the training samples), we propose to match the distribution of $\mathbf{z}^{CF}$ to $\mathbf{z}$ in an adversarial training way. We train a discriminator $f^{dis}(\cdot)$ to classify $\mathbf{z}^{CF}$ and $\mathbf{z}$. The objective of $f^{mod}(\cdot)$ is to fool the discriminator by minimizing the loss function:
\begin{equation}
	\label{adversal}
	\mathcal{L}_{adv} = \sum^{N}_{i=1}-\texttt{log} (f^{dis}(\mathbf{z}_i^{CF})).
\end{equation} 
Conversely, the objective of the discriminator is to minimize the loss $\sum^{N}_{i=1}-(\texttt{log} (f^{dis}(\mathbf{z}_i)) + \texttt{log}(1 - f^{dis}(\mathbf{z}_i^{CF}))$. During the training stage, we iteratively optimize the modulation network $f^{mod}(\cdot)$ and the discriminator $f^{dis}(\cdot)$. This is similar to the training procedure of AAE \cite{makhzani2015adversarial}. While AAE matches the latent space to an assumed distribution (e.g. standard Gaussian distribution), we match the latent distribution of counterfactuals to the latent distribution of the training samples.

The above three optimisation criteria: the classification loss, the nearest loss and the adversarial loss, can be combined into the final objective function: 
\begin{equation}
	\label{finalloss}
	\begin{aligned}
	&\mathcal{L}_{total} = \sum^{N}_{i=1} \Big( \alpha_1 \text{HingeLoss}(h(\mathbf{x}_i^{CF}), y_i^{CF}, \beta) +
	\alpha_2\big(\text{dist}(\mathbf{x}_i, \mathbf{x}_i^{CF})
	+ ||\boldsymbol{\delta}_i||^2_2\big)
	+ \alpha_3 \big( -\log\ (f^{dis}(\mathbf{z}_i^{CF}))\big)\Big)
	\end{aligned}
\end{equation}
where $\alpha_1, \alpha_2, \alpha_3$ are hyperparameters to balance the effects of the three criteria.

\section{Experiments}
We conduct our method on both synthetic and real-world datasets to answer the following research questions:

\begin{itemize}[leftmargin=*]
	\item \textbf{RQ1}: How effective is our method in preserving the relationships when generating counterfactuals?
	\item \textbf{RQ2}: How can our method provide suggestions in a real-world machine learning application?
\end{itemize}

\subsection{Experiment Settings}
\noindent\textbf{Datasets.} We evaluate our approach on two synthetic datasets including a toy dataset and a complex nonlinear dataset, a publicly-available real-world simulated Bayesian network dataset Sangiovese\footnote{\href{https://www.bnlearn.com/bnrepository/clgaussian-small.html}{https://www.bnlearn.com/bnrepository/clgaussian-small.html}} and a real-world diabetes dataset Pima-Indians-Diabetes\footnote{\href{https://www.kaggle.com/kumargh/pimaindiansdiabetescsv}{https://www.kaggle.com/kumargh/pimaindiansdiabetescsv}}. The statistics of the datasets are shown in Table \ref{tab:statics}.

\begin{table}[t!]
	\centering
	\caption{Statistics of datasets. Splits mean the splits of train/val/test.}
	\scalebox{1}{
	\begin{tabular}{lcccccccc}
	\toprule
	Datasets &Synthetic-toy &Synthetic-nonlinear &Sangiovese& Diabetes
	\cr
	\midrule
	\# Classes & 2 & 2 & 2 & 2 \cr
	\# Samples & 20000 & 20000 & 11570 & 336 \cr
    Train/val/test Splits  & 8:1:1 & 8:1:1 & 8:1:1 & Leave-out-one\cr
	\bottomrule
	\end{tabular}}
	\vspace{-2pt}
	\label{tab:statics}
\end{table}

\noindent\textbf{Compared methods.} We implement the example-based Plain-CF method that was proposed in \cite{wachter2018counterfactual} with Hinge loss and $L_2$ norm. Following \cite{dandl2020multi}, we further implement another version of Plain-CF$_{K}$ that adds an extra constraint on the distances to $k$-nearest neighbours. We also compare our method to the CF-VAE proposed in \cite{mahajan2019preserving}. This is a base model that does not consider possible relationships between the predictors. To be consistent with the experimental setting in \cite{mahajan2019preserving} which manually label 10\% of the counterfactuals, we compare the example-based VAE (EB-VAE) by manually labeling 10\% of the counterfactuals and retrain the CF-VAE.

\noindent\textbf{Evaluation metrics.} The proposed approach is evaluated both quantitatively and qualitatively. For the quantitative metrics, we :
\begin{itemize}[leftmargin=*]
	\item \textit{Validity} that is defined as a percentage of counterfactuals that are generated with correct labels.
	\item \textit{Constraint Feasibility Score} $S$ that is the percentage of counterfactuals that preserve the relationships \cite{mahajan2019preserving, downs2020cruds}. For example, the constraint feasibility score is 90\% if 90\% of the counterfactuals have two positively related attributes both increase or both decrease. For two relationships, the score is calculated as $2 \frac{S_1 S_2}{S_1 + S_2}$ where $S_1$ and $S_2$ are the scores for the two relationships separately.
	\item \textit{Euclidean Distance} measures the distance between counterfactuals and corresponding original samples. The distance is normalized by the diameter of the set of original samples. This metric measures how much the original sample should be perturbed to obtain a corresponding counterfactual.
	\item \textit{Mahalanobis Distance} is another metric that accounts for covariance in the data. The distance between two points is defined as $D(\mathbf{x_1}, \mathbf{x_2}) = \sqrt{(\mathbf{x_1}-\mathbf{x_2})^\top\Sigma^{-1}(\mathbf{x_1} - \mathbf{x_2})}$ where $\Sigma$ denotes the covariance matrix.
	
\end{itemize}

It is worth noting that the distances as evaluation metrics might not accurately reflect the quality of generated counterfactuals. There are cases when in order to preserve the relationships between the attributes, a larger perturbation is required and hence results in larger distances. Figure \ref{fig:expl} demonstrates an example of a dataset with attributes that exhibit a linear relationship, a generated counterfactual marked in yellow is closer in terms of Euclidean distance than the one marked in green, however such counterfactual does not account for linear relationship between the attributes, while the green counterfactual preserves the relationship. The Mahalanobis distance on the other hand would produce a smaller distance for the green counterfactual as it accounts for the data covariance.

The qualitative evaluations include visualization of selected pairs of attributes and their corresponding counterfactuals, as well as, $t$-SNE visualization of the complete attribute space. To visually verify that the relationships are preserved, we plot arrows connecting original samples with their corresponding counterfactuals.

\noindent\textbf{Details on implementation.} The models employed as black-box classifiers are two-layer MLPs with ReLU activation functions. We set the dimensions of the hidden layer to 32 for synthetic datasets and 16 for the Sangiovese dataset. The test accuracy of the two synthetic datasets, were nearly 98\%, while for Sangiovese it was around 83\%. We used the official Pytorch implementation of DAG-GNN\footnote{\href{https://github.com/fishmoon1234/DAG-GNN}{https://github.com/fishmoon1234/DAG-GNN}}, both the encoder and the decoder are two layers MLP with hidden size of 16 neurons, and the latent size of 4 neurons (we did not tune these two hyperparameters). Both the modulation network and the discriminator are two-layer MLPs with equal numbers of hidden units. For synthetic datasets, the modulation network was trained for 100 epochs while for Sangiovese, it was trained for 80 epochs. Grid search was used to tune the hyperparameters of the two networks where the hidden sizes were chosen from $[\texttt{32, 64}]$, the learning rates were chosen from $[\texttt{5e-4, 1e-3, 2e-3}]$ and the batch sizes were chosen from $[\texttt{16, 32}]$. The choice of hyperparameters did not significantly affect the validity that reached nearly $100\%$, and hence we chose the hyperparameters to maximise constraint feasibility score. The hyperparameters $\alpha_1$, $\alpha_2$ and $\alpha_3$ were set to $0.5$, $1$, and $0.5$, respectively. The codes are available at \href{https://github.com/xintao-xiang/Realistic_CF/tree/multi_label}{https://github.com/xintao-xiang/Realistic\_CF/tree/multi\_label}.

\subsection{Performances Comparison (RQ1)}

\noindent\textbf{Synthetic 5-variable Toy Dataset.}  The purpose of this experiment is to evaluate the proposed model on a toy dataset with linear relationships between the attributes. The data is generated using the following structural equations and noise distributions:
\begin{alignat}{2}
X_1 &= U_1  &\quad\quad\quad\quad U_1 &\sim \mathcal{N}(0, 0.5) \\  
X_2 &= U_2  & U_2 &\sim \mathcal{N}(0, 0.5) \\  
X_3 &= 2X_1 - X_2 + U_3 & U_3 &\sim \mathcal{N}(0, 0.5) \\
X_4 &= -2X_3 + U_4 & U_4 &\sim \mathcal{N}(0, 0.5) \\
X_5 &= \texttt{sin}(X_3) + U_5 & U_5 &\sim \mathcal{N}(0, 0.5)
\end{alignat}
The dataset contains 20000 samples, and is split into training, validation and testing sets following the  8:1:1 ratio. The samples are labeled to 1 when $\texttt{sin}(X_i) > 0.5$ for more than 2 variables (without added Gaussian noise). For this synthetic data, we aim at capturing the relationships between $X_3$ and $X_4$. \textit{If $X_3$ decrease/increase, $X_4$ increase/decrease}.

\begin{figure}[t!]
    \centering
	\begin{tabular}{@{}c@{}c@{}c@{}}
	  \includegraphics[width=.4\textwidth, trim={0cm 0.2cm 0cm 0.2cm}, clip]{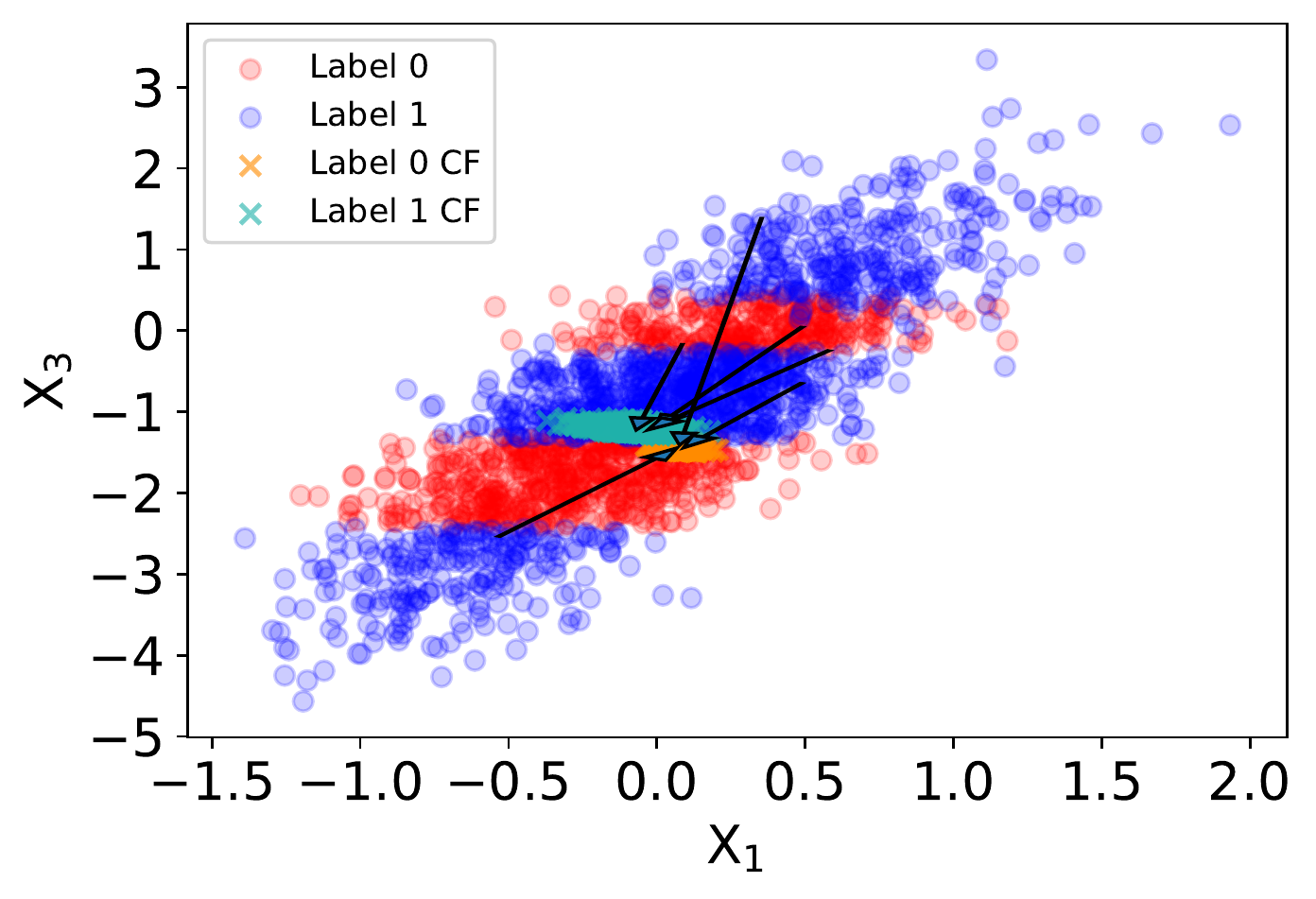} &
	  \includegraphics[width=.4\textwidth, trim={0cm 0.2cm 0cm 0.2cm}, clip]{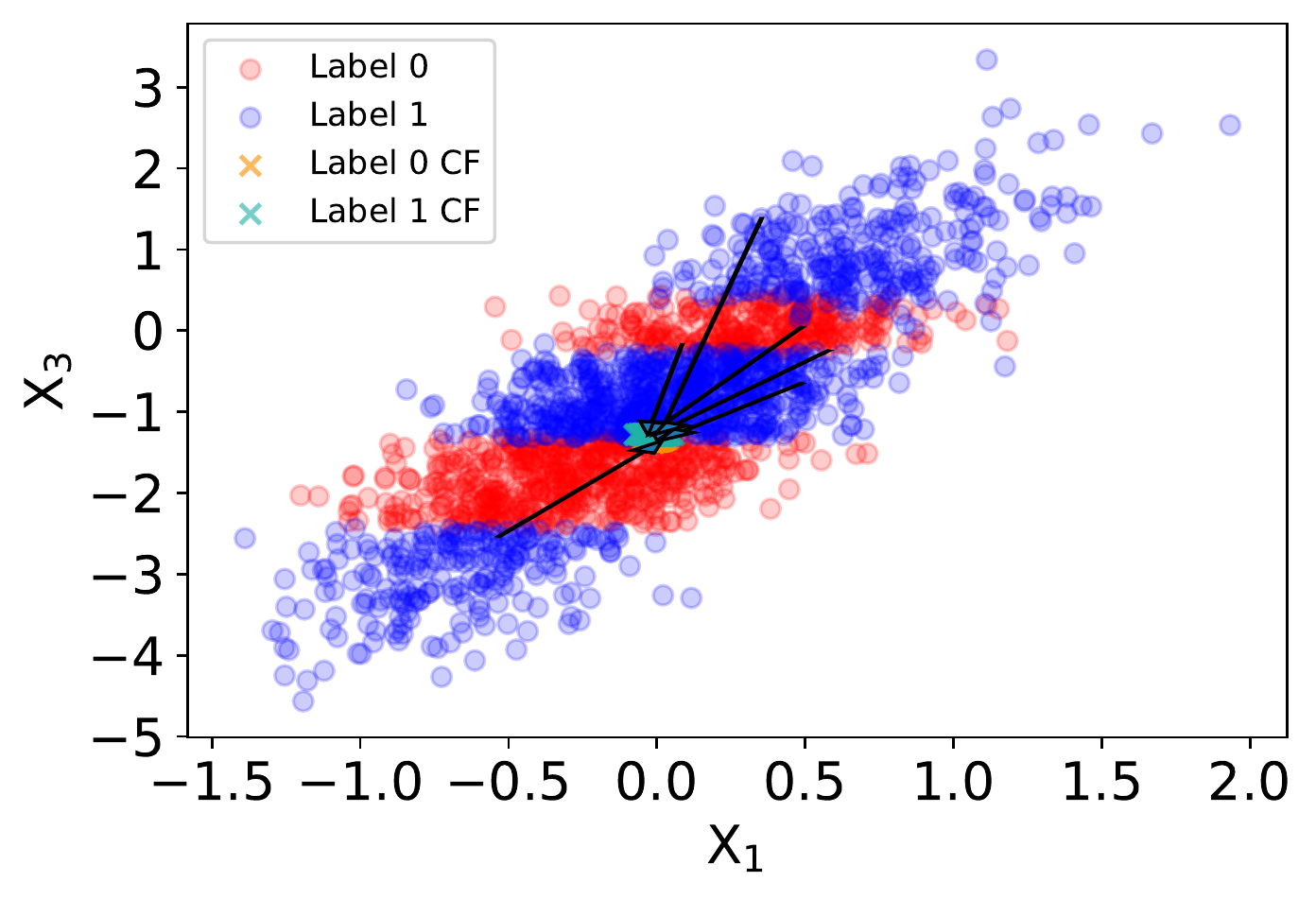} \\
	  \specialrule{0em}{-3pt}{0pt}
	  \footnotesize (a) CF-VAE &\footnotesize (b) EB-VAE
	\end{tabular}
	
	\begin{tabular}{@{}c@{}c@{}}
		\includegraphics[width=.4\textwidth, trim={0cm 0.2cm 0cm 0.2cm}, clip]{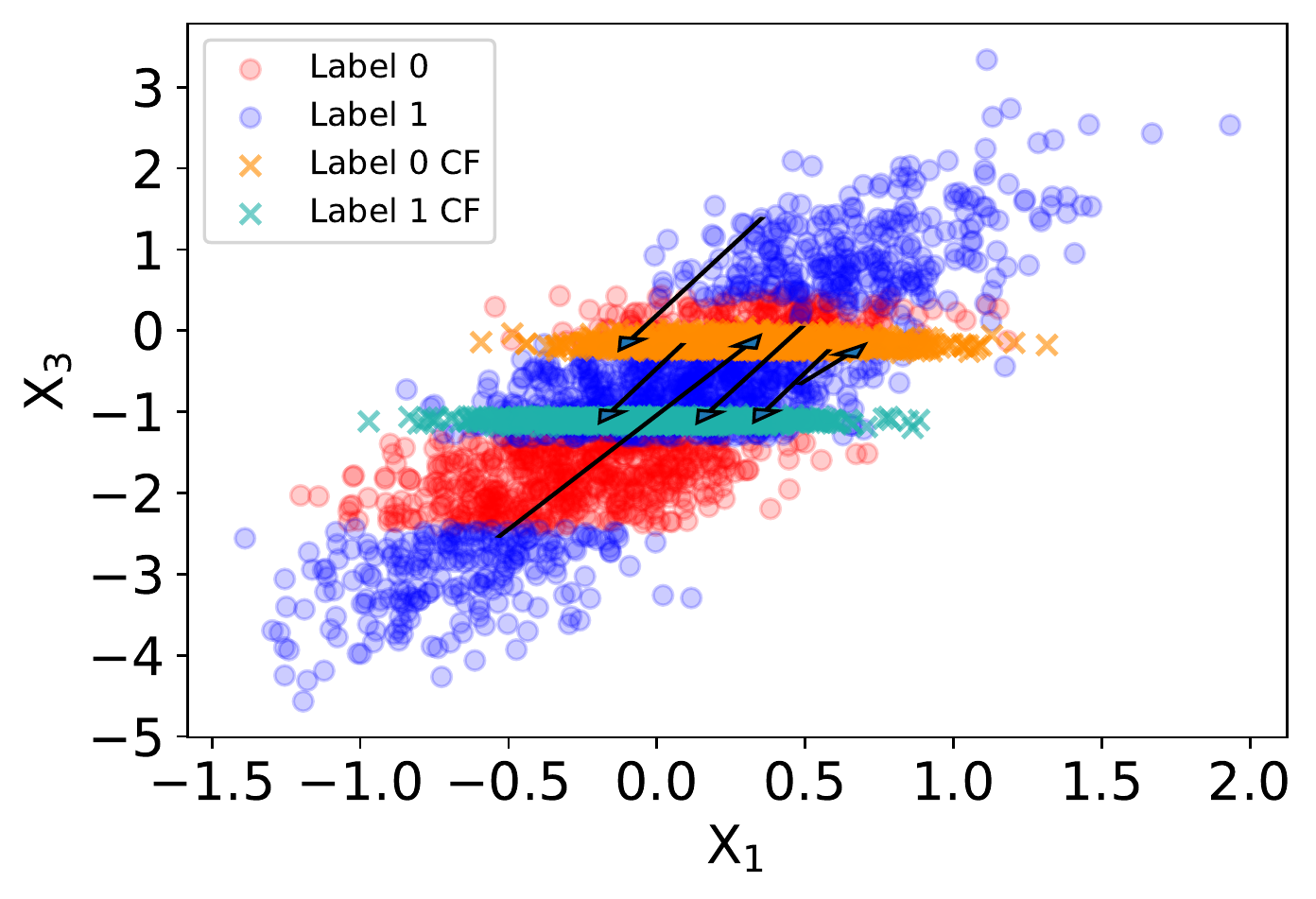}  \\
		\specialrule{0em}{-3pt}{-1pt}
		\footnotesize (c) Ours
	\end{tabular}
	
    \caption{Visualization of related variables of the synthetic toy dataset. The arrows indicate the directions from true samples to counterfactuals. CF means the generated counterfactuals. }\label{synthetic_toy_vis}
    
\end{figure}

\begin{figure*}[t!]
	\centering
	\begin{tabular}{ccc}
		\includegraphics[width=.32\textwidth, trim={0.9cm 1.5cm 1.5cm 1.2cm}, clip]{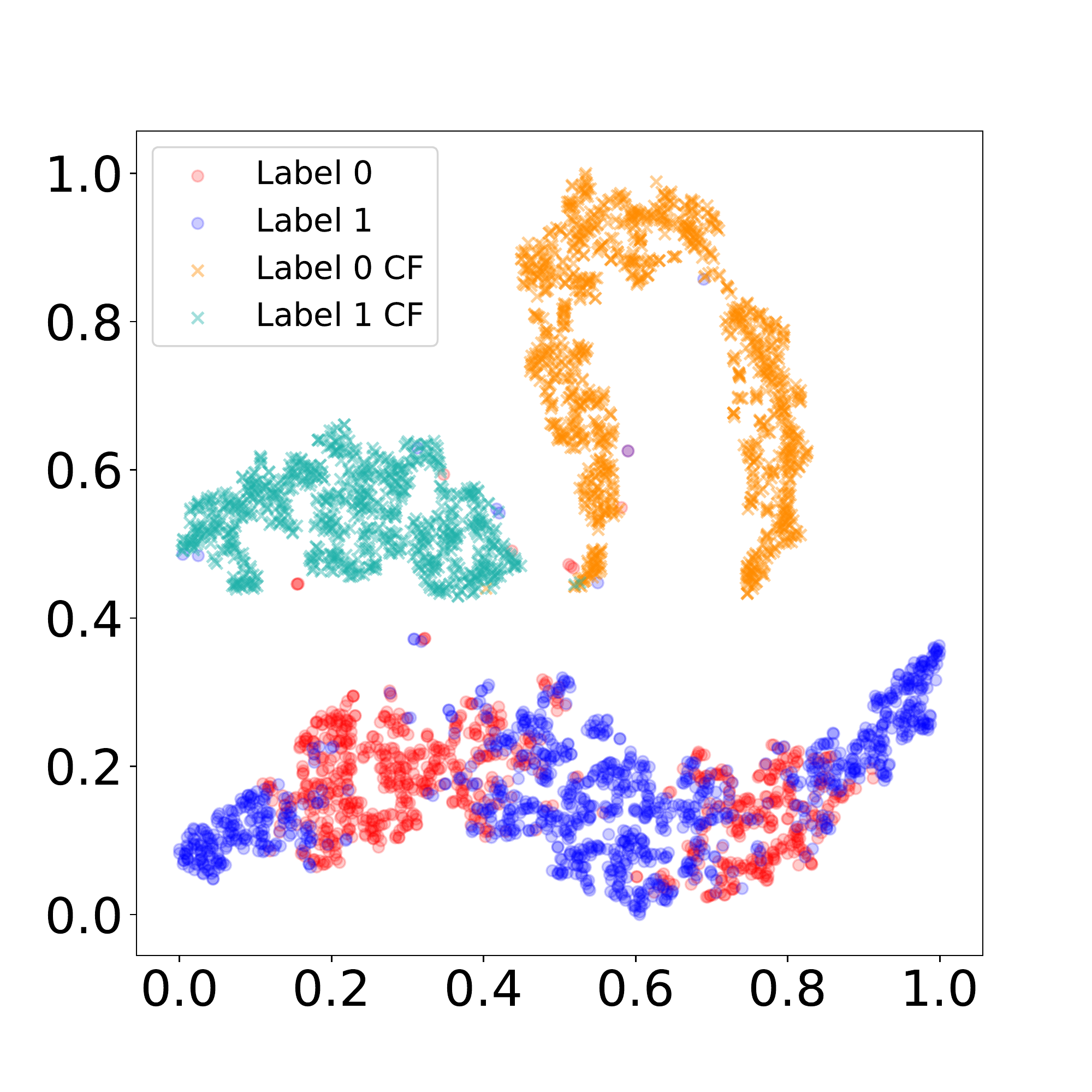} &
		\includegraphics[width=.32\textwidth, trim={0.9cm 1.5cm 1.5cm 1.2cm}, clip]{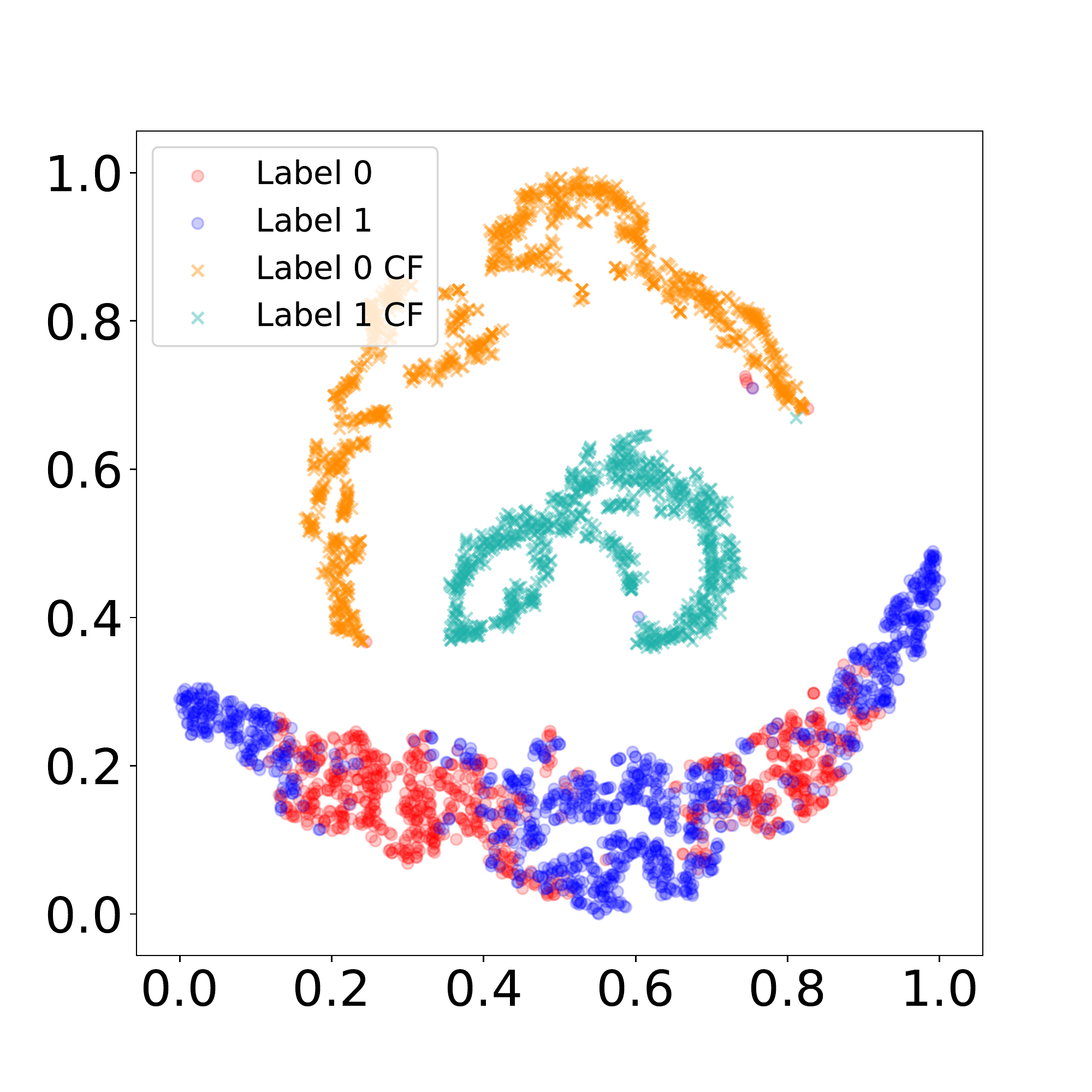} & \includegraphics[width=.32\textwidth, trim={0.9cm 1.5cm 1.5cm 1.2cm}, clip]{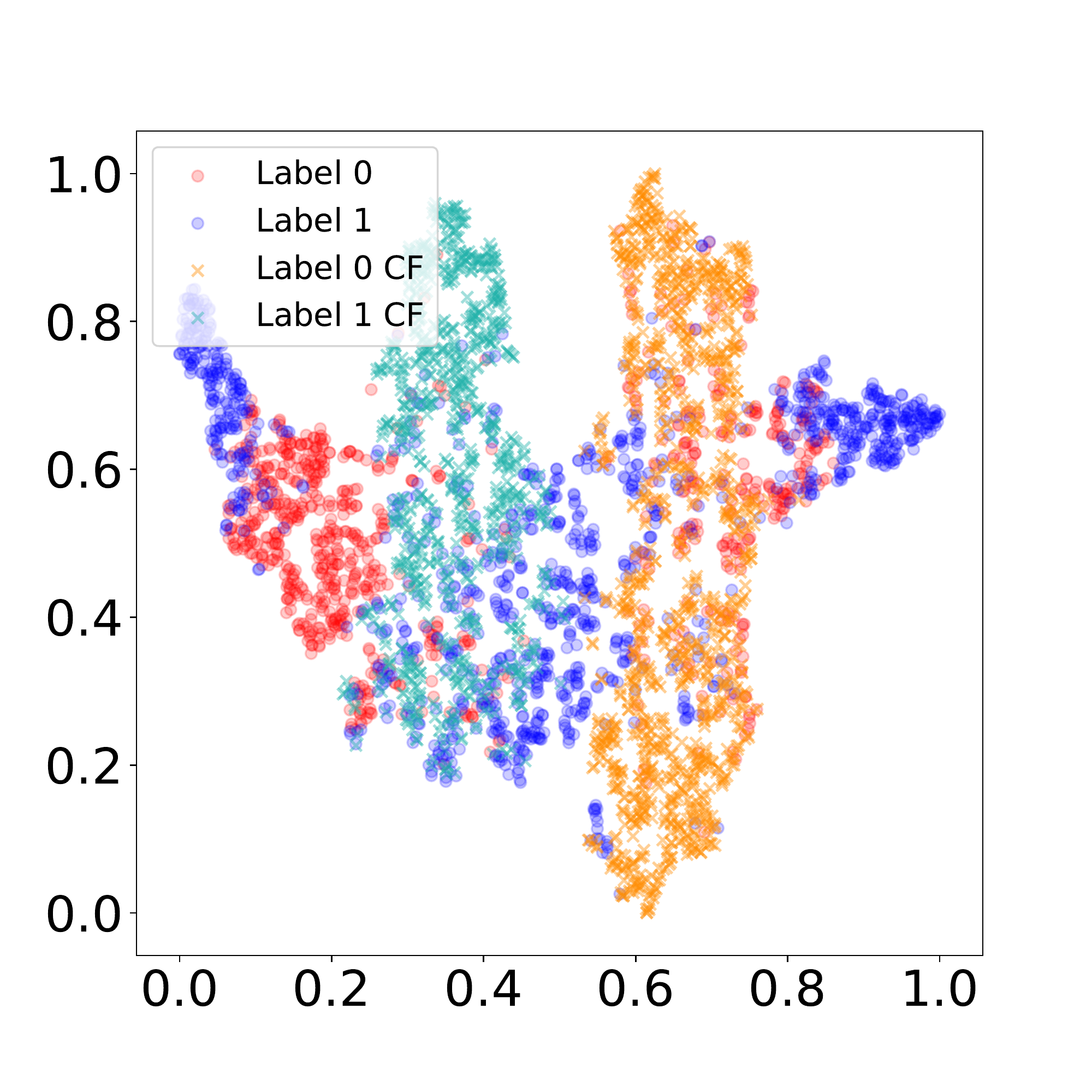}\\
		\footnotesize (a) CF-VAE &\footnotesize (b) EB-VAE & \footnotesize (c) Ours
	\end{tabular}

	\caption{$t$-SNE visualization of synthetic toy dataset.}
	\label{synthetic_toy_tsne}
\end{figure*}

The results are shown in the left half of table \ref{tab:performance1}. Nearly all models generated 100\% valid counterfactuals. In terms of the constraint feasibility score, the proposed approach is among the best. Our model achieved the smallest Euclidean distance. Figure \ref{synthetic_toy_vis} demonstrates that the counterfactuals generated by EB-VAE model tend to cluster to a single point. Such counterfactuals, due to a lack of diversity, are hardly acceptable in practice. Besides, by comparing the directions of the difference vectors between the original and the corresponding counterfactuals, it is easy to see that the vectors of the proposed method align with the true linear relationship existing in the data. Further, the counterfactuals generated by the proposed method exhibits a larger visual overlap with the data distribution of the corresponding class. Figure \ref{synthetic_toy_tsne} provides further intuition on the data distributions projected by $t$-SNE to a two dimensional space. The counterfactuals and the samples of the corresponding class have a significant overlap while preserving clear boundaries with the other class. These figures qualitatively support the superiority of the proposed approach in generating realistic counterfactuals.

\begin{table*}[t!]
	\centering
	\caption{Experimental results on synthetic data}
	\scalebox{0.85}{
	\begin{tabular}{lcccccccc}
	\toprule
	\multirow{2}{*}{Methods}&
	\multicolumn{4}{c}{Synthetic Toy}&
	\multicolumn{4}{c}{Synthetic Non-linear}\\
	\cmidrule(r){2-5}\cmidrule(l){6-9}
	&Valid (\%)&Const (\%)&Euclidean Dist& Mahalanobis Dist &Valid&Const (\%)&Euclidean Dist & Mahalanobis Dist
	\cr
	\midrule
	Plain-CF  & 97.25 & 51.50 & 0.1490 & 6.6867 & 92.20 & 55.34 & 0.1884  & 4.8070  \cr
	Plain-CF$_{K}$  & 99.45 & 58.85 &  0.1195 & 5.7732  & 98.05 & 64.65 & 0.2148 & 4.4680  \cr
	CF-VAE  & 100 & 74.36 & 0.1404  & 2.9336  & 100 & 67.85 & 0.1879 & 3.3138\cr
	EB-VAE  & 100 & 77.66 & 0.1421  & 2.5013 & 100 & 67.19 & 0.1867 & 3.0320\cr
	Ours  & 99.73 & 93.00 & 0.1474 & 2.3493 & 99.99 & 84.4  & 0.2263 & 3.4418\cr

	\bottomrule
	\end{tabular}}
	\label{tab:performance1}
\end{table*}

\begin{figure}[t!]
	\centering
	\begin{tabular}{@{}c@{}c@{}}
	  \includegraphics[width=.4\textwidth, trim={0.25cm 0.2cm 0.25cm 0cm}, clip]{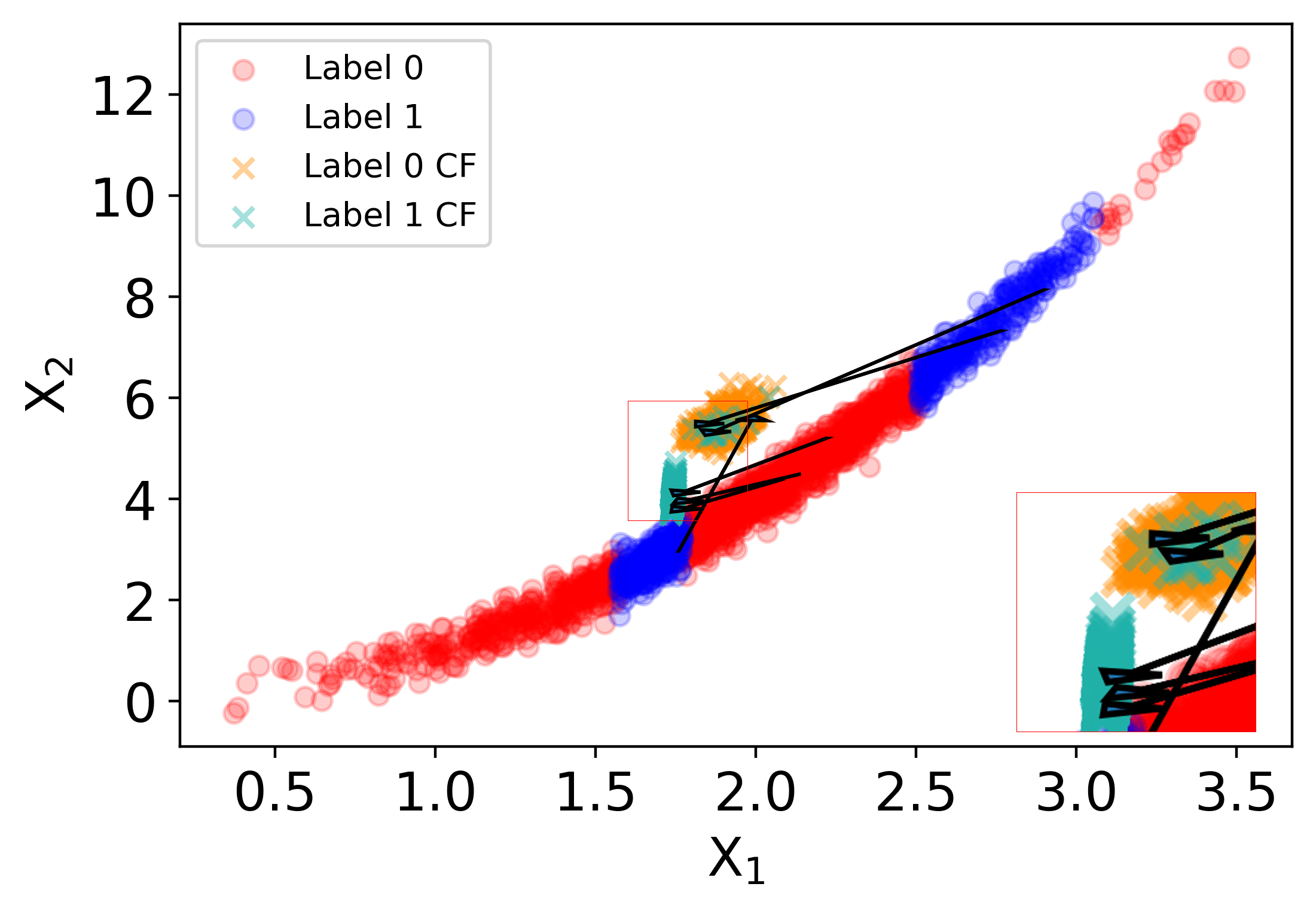}
	  \includegraphics[width=.4\textwidth, trim={0.25cm 0.2cm 0.25cm 0cm}, clip]{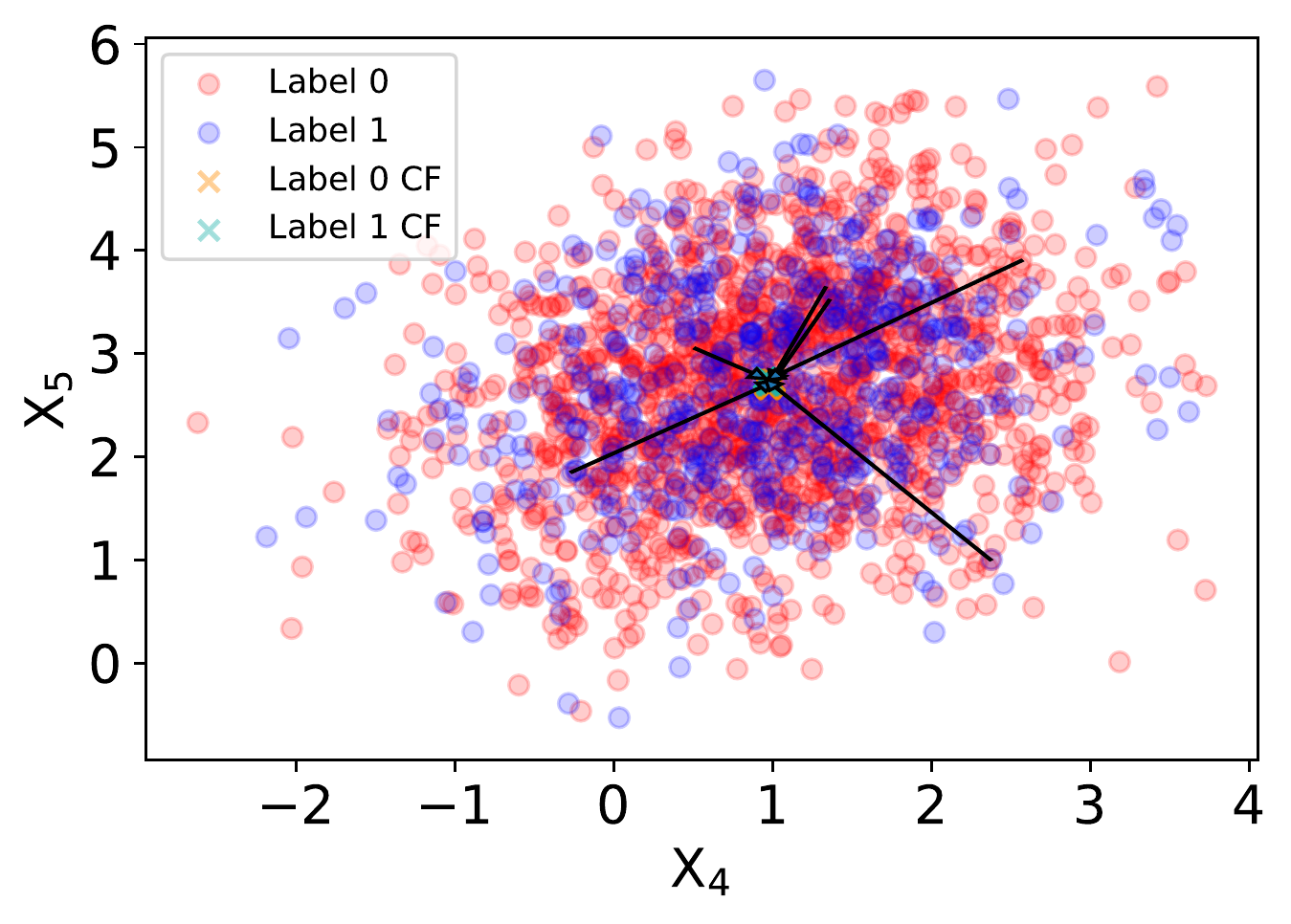}\\
	  \specialrule{0em}{-5pt}{-1pt}
	  \footnotesize (a) CF-VAE 
	\end{tabular}
 
	\begin{tabular}{@{}c@{}c@{}}
		\includegraphics[width=.4\textwidth, trim={0.25cm 0.2cm 0.25cm 0cm}, clip]{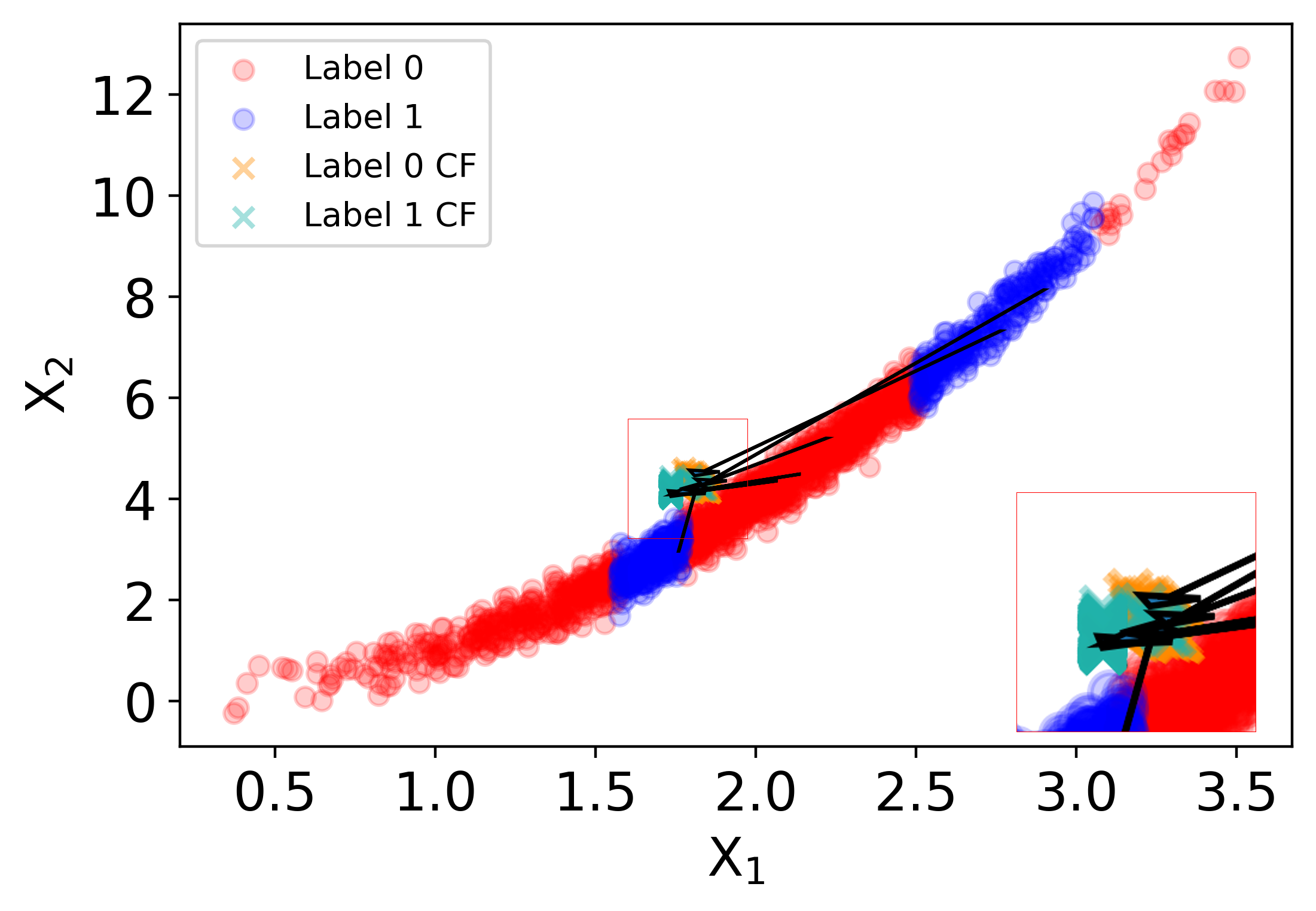}
		\includegraphics[width=.4\textwidth, trim={0.25cm 0.2cm 0.25cm 0cm}, clip]{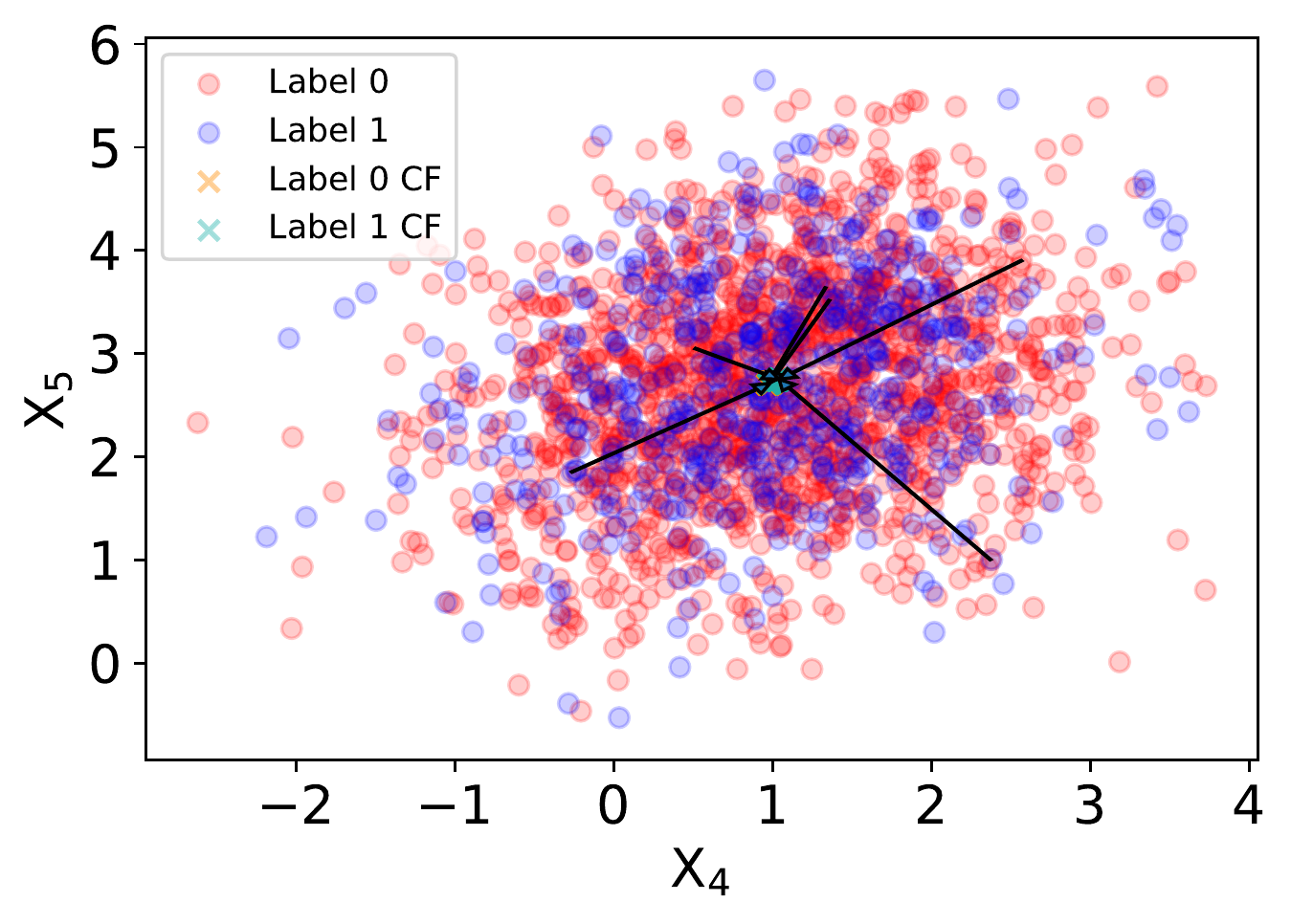}\\
		\specialrule{0em}{-5pt}{-1pt}
		\footnotesize (b) EB-VAE 
	\end{tabular}
	
	\begin{tabular}{@{}c@{}c@{}}
        \includegraphics[width=.4\textwidth, trim={0.25cm 0.2cm 0.25cm 0cm}, clip]{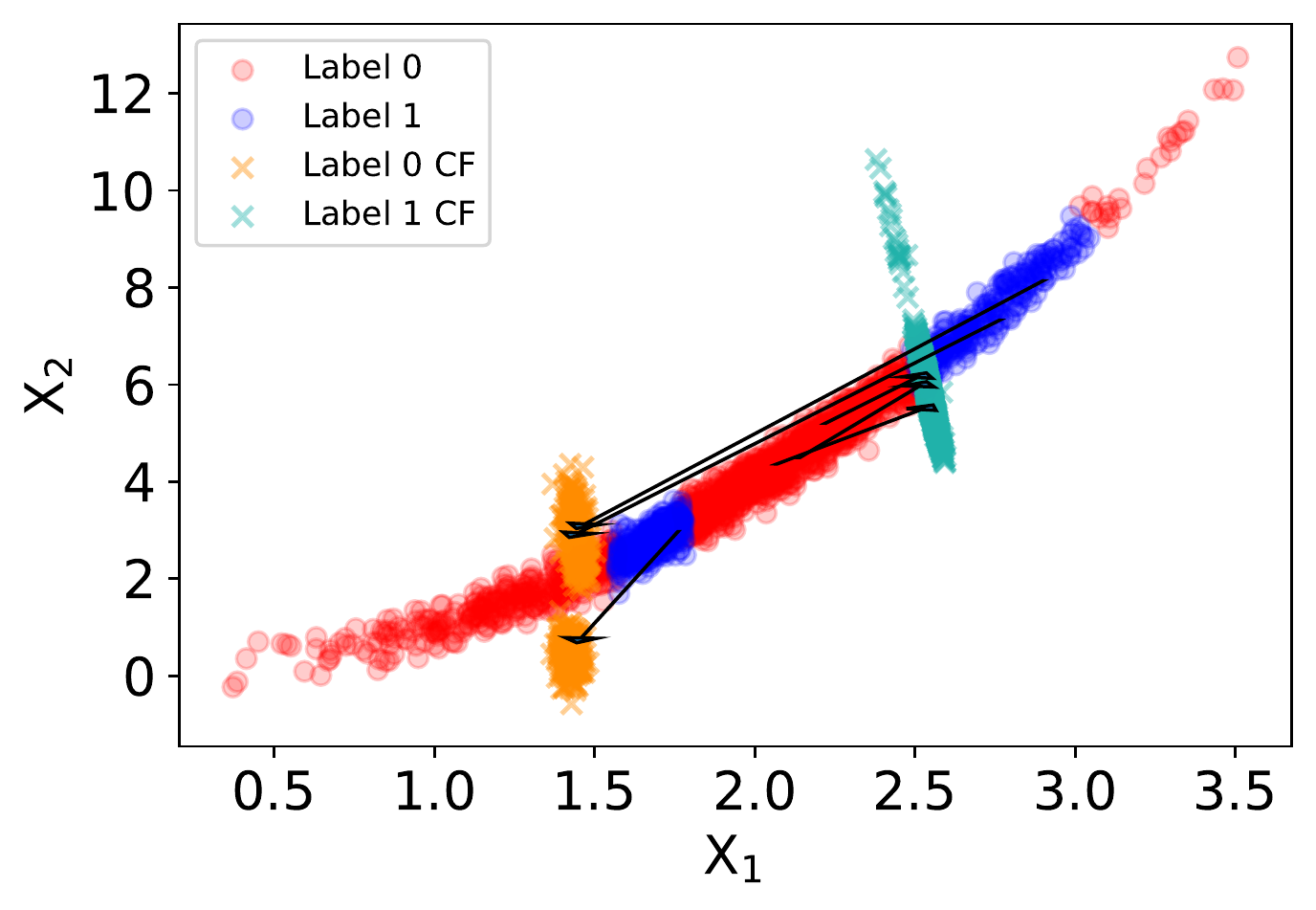}
		\includegraphics[width=.4\textwidth, trim={0.25cm 0.2cm 0.25cm 0cm}, clip]{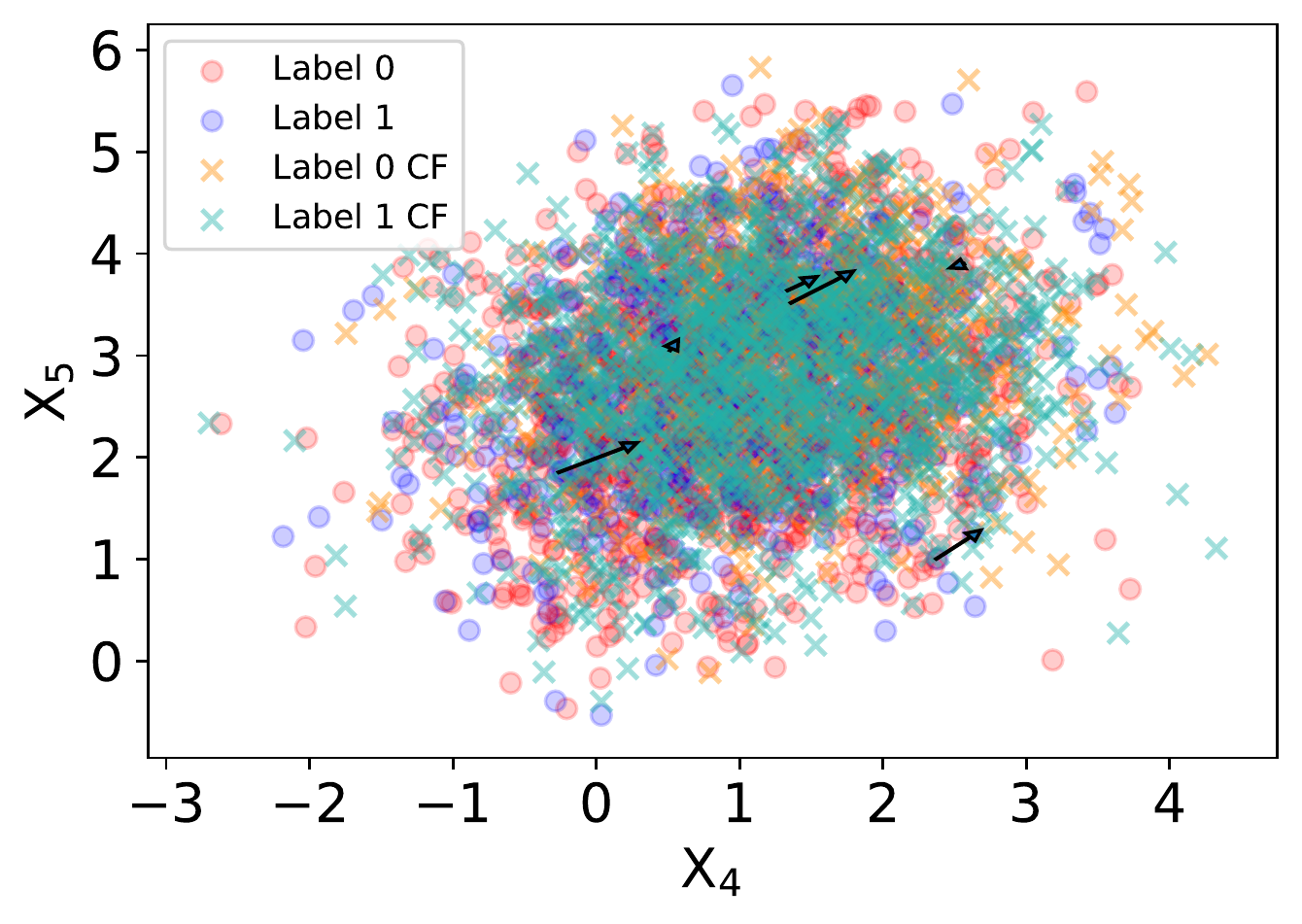}\\
		\specialrule{0em}{-5pt}{-1pt}
		\footnotesize (c) Ours
	\end{tabular}
	\caption{The visualization related variables in synthetic non-linear dataset.}
	\label{synthetic_nonlinear_vis}
\end{figure}

\noindent\textbf{Synthetic 5-variable Non-linear Dataset.} This dataset includes more complex and non-linear relationships between variables. The SEM are as follows:
\begin{alignat}{2}
X_1 &= U_1  &\quad\quad\quad U_1 &\sim \mathcal{N}(2, 0.5) \\  
X_2 &= X_1^2 + U_2  & U_2 &\sim \mathcal{N}(0, 0.25) \\  
X_3 &= \texttt{sin}(-2X_1) + U_3 & U_3 &\sim \mathcal{N}(0, 1) \\
X_4 &= U_4 & U_4 &\sim \mathcal{N}(1, 1) \\
X_5 &= \frac{1}{\texttt{e}^{-1.5X_4}} + 2 + U_5 & U_5 &\sim \mathcal{N}(0, 1)
\end{alignat}
The dataset contains 20000 samples, and is split into training, validation and testing sets following the  8:1:1 ratio. The samples are labeled to 1 when $\texttt{sin}(X_i) > 0$ for all the 5 variables (without added Gaussian noise). This dataset includes two relationship constraints: \textit{(1) If absolute values of $X_1$ increase/decrease, $X_2$ increase/decrease, and (2) if $X_4$ increase/decrease, $X_5$ increase/decrease}. 

Summarized results are presented in the right half of Table \ref{tab:performance1}. The results are similar to that of synthetic toy dataset. All the models except the Plain-CF generate counterfactuals with almost 100\% valid labels. Our method has the highest constraint feasibility score indicating that our method preserves the relationships between the attributes. The distributions of the generated and the original samples are shown in Figure \ref{synthetic_nonlinear_vis}. Compared to other models, the counterfactuals generated by the proposed model has a larger intersection with the original sample distributions of the corresponding class. By comparing the arrows indicating the direction of the changes between the original samples and their corresponding counterfactuals, we observe that for the baseline methods, the samples tend to cluster into a small group and ignore the relationships. This is especially true for the pair $X_4$ and $X_5$. Our method, on the other hand, preserves the relationships. Having said that, all methods fail in capturing the non-linear relationships, however our method performs better than the compared methods.

\begin{table}[t!]
	\centering
	\caption{Experimental results on Sangiovese.}
	\vspace{-2pt}
	\scalebox{0.9}{
	\begin{tabular}{lcccccccc}
	\toprule
	Methods &Valid (\%)&Const (\%)&Euclidean Dist& Mahalanobis Dist
	\cr
	\midrule
	Plain-CF  & 100 & 30.52 & 0.2821 & 6.2232 \cr
	Plain-CF$_{K}$  & 100 & 63.17 & 0.2665& 5.0929 \cr
	CF-VAE  & 99.84 & 63.03 & 0.2169 & 4.7695 \cr
	EB-VAE  & 99.97 & 67.19 & 0.1857 & 3.6738 \cr
	Ours  & 99.98 & 99.69 & 0.1961 & 1.9971 \cr

	\bottomrule
	\end{tabular}}
	\label{tab:performance2}
\end{table}

\begin{figure}[t!]
	\centering
	\begin{tabular}{@{}c@{}c@{}}
	  \includegraphics[width=.4\textwidth, trim={0.25cm 0.2cm 0.25cm 0.25cm}, clip]{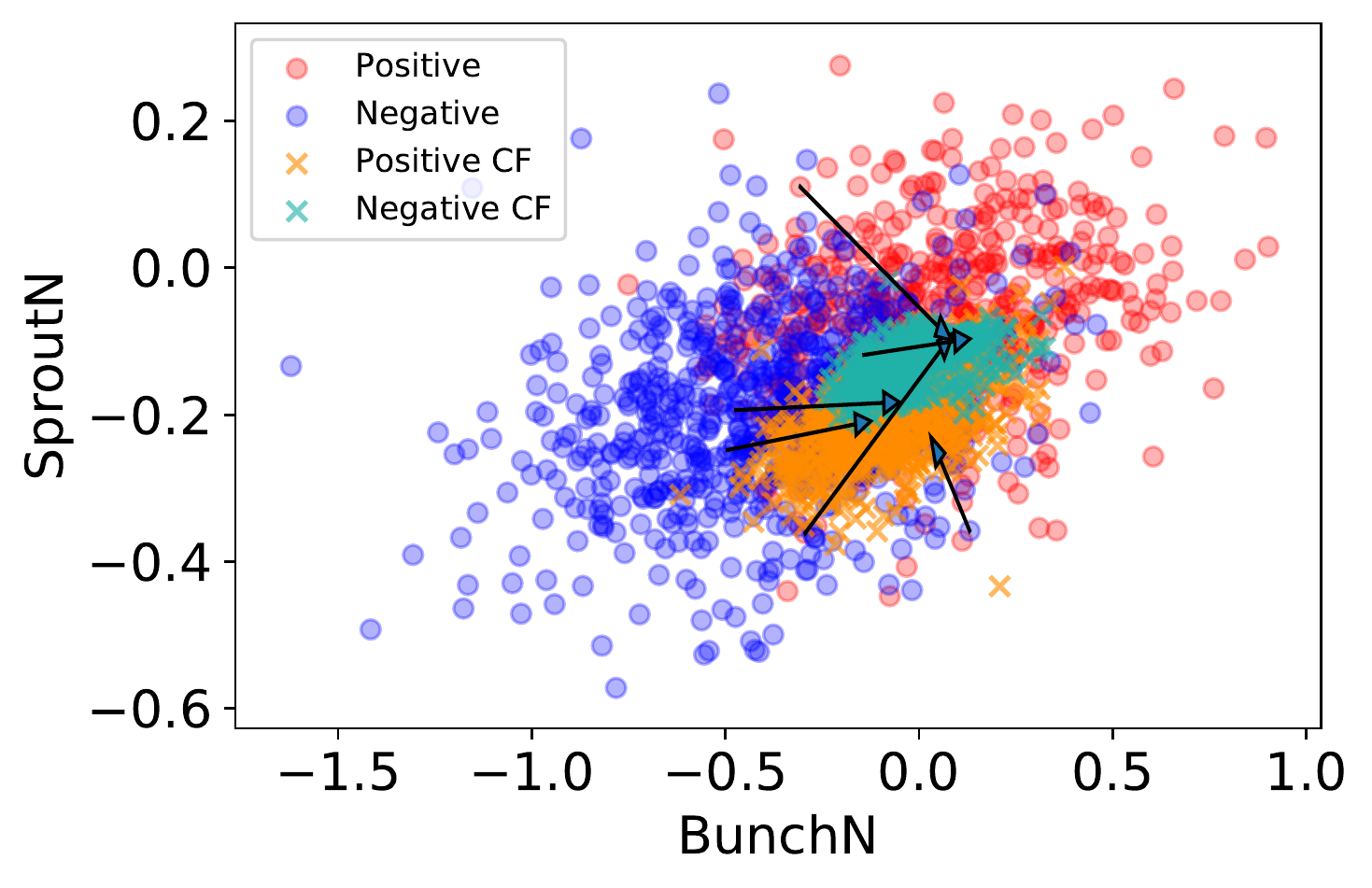} 
	  \includegraphics[width=.4\textwidth, trim={0.25cm 0.2cm 0.25cm 0.25cm}, clip]{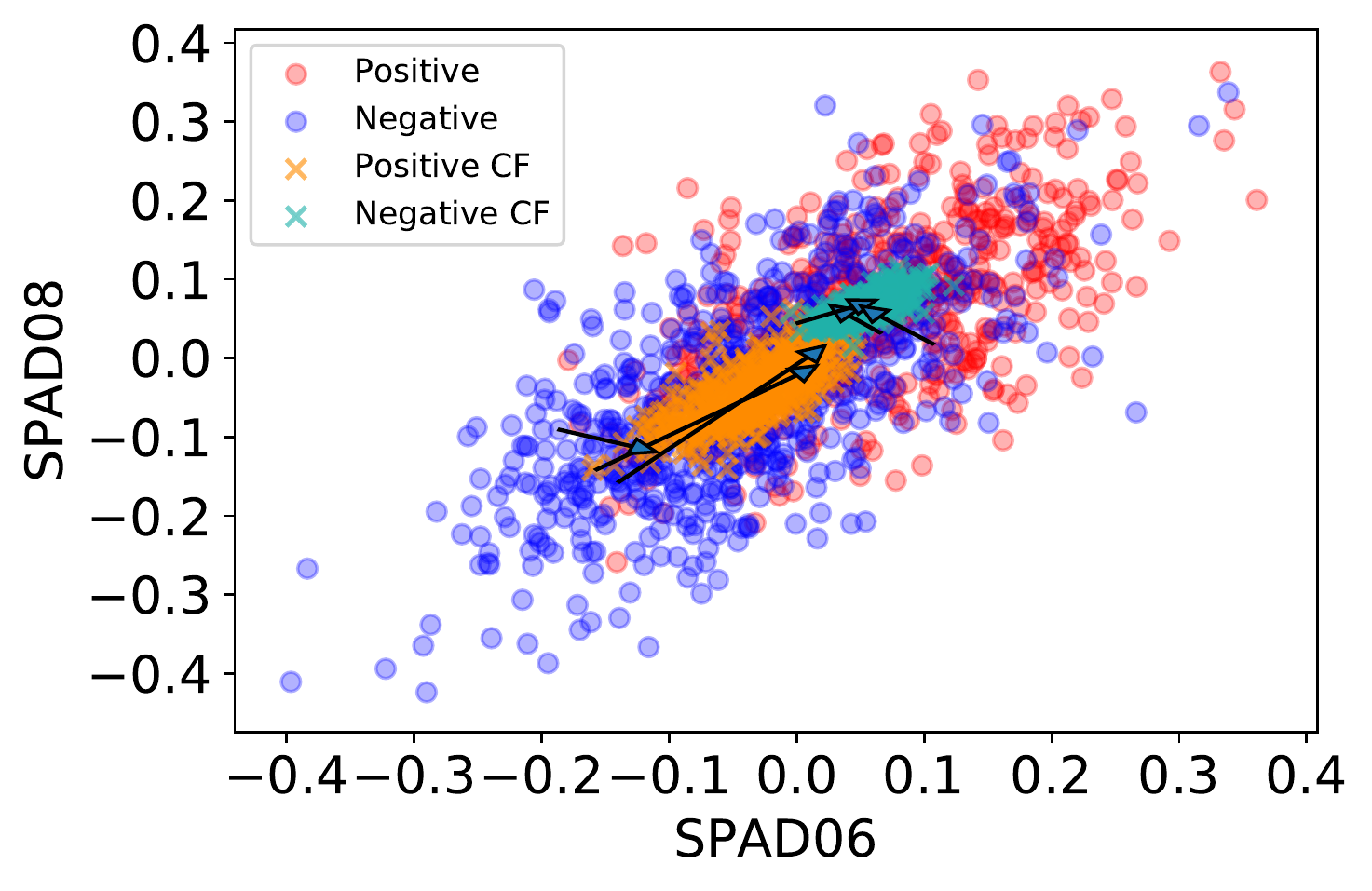} \\ \specialrule{0em}{-5pt}{-1pt}
	  \footnotesize (a) CF-VAE 
	\end{tabular}
    
	\begin{tabular}{@{}c@{}c@{}}
	  \includegraphics[width=.4\textwidth, trim={0.25cm 0.2cm 0.25cm 0.25cm}, clip]{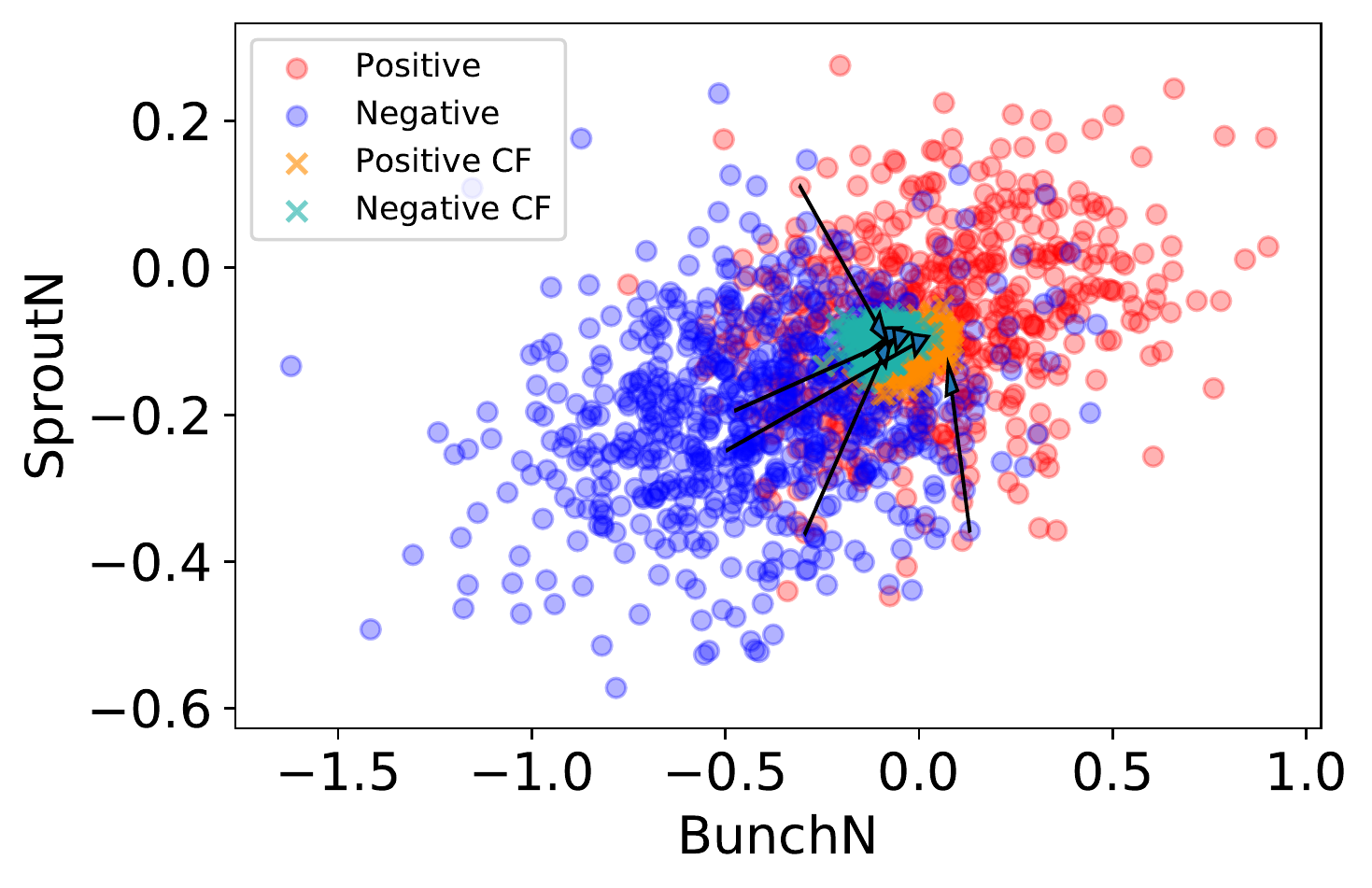} 
	  \includegraphics[width=.4\textwidth, trim={0.25cm 0.2cm 0.25cm 0.25cm}, clip]{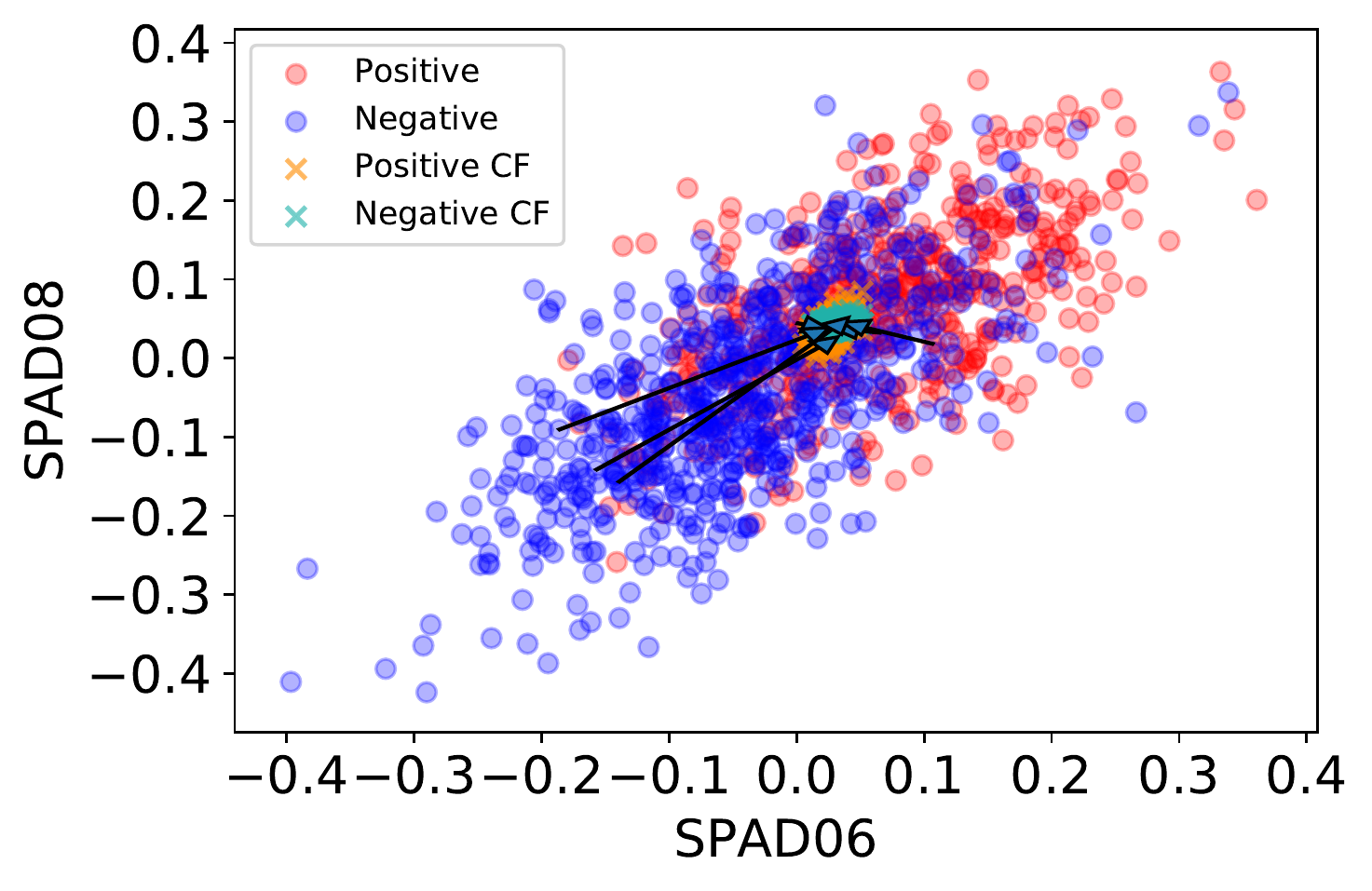} \\
        \specialrule{0em}{-5pt}{-1pt}	   
    \footnotesize (b) EB-VAE
	\end{tabular}

	\begin{tabular}{@{}c@{}c@{}}
	 \includegraphics[width=.4\textwidth, trim={0.25cm 0.2cm 0.25cm 0.25cm}, clip]{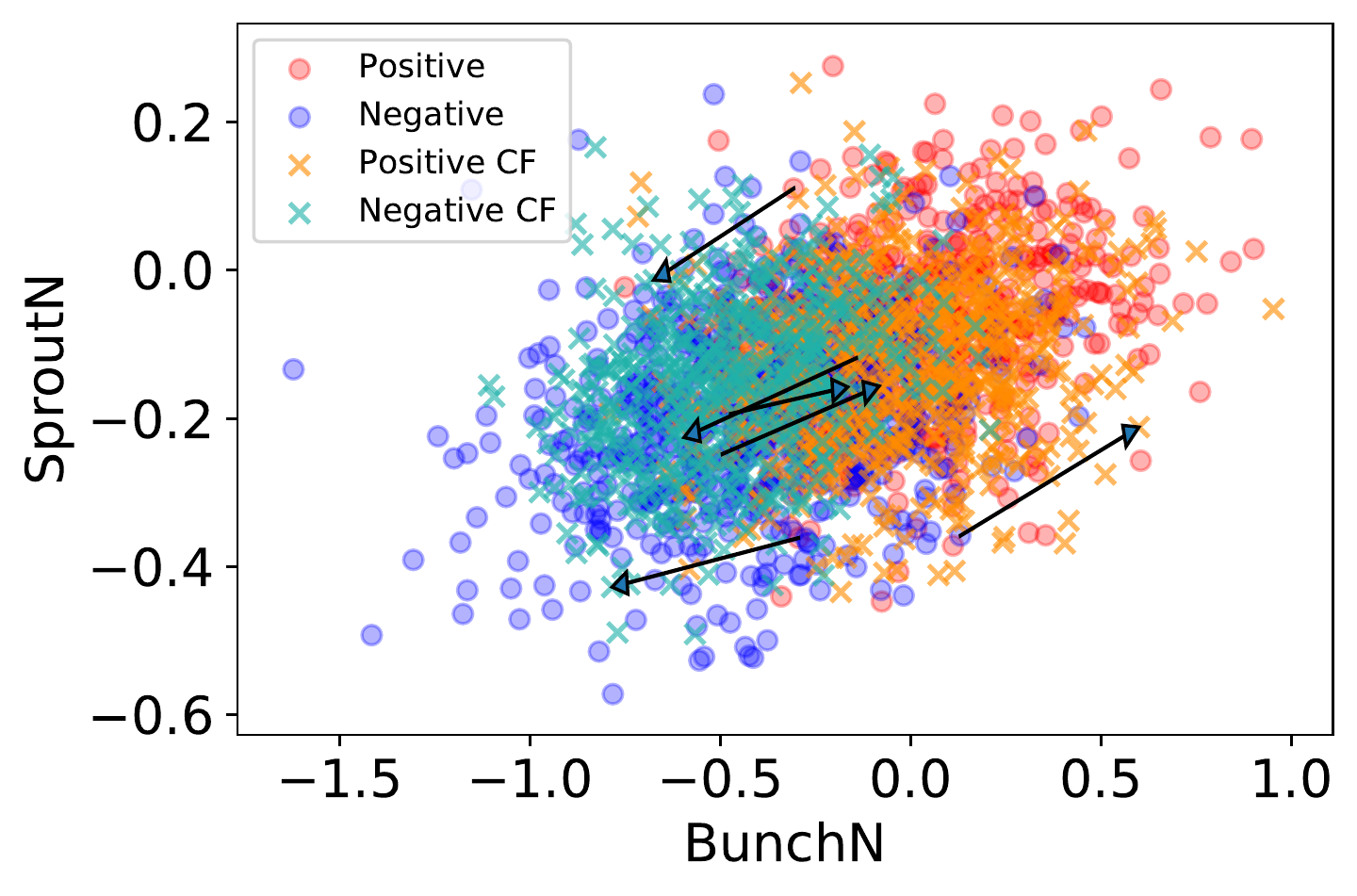}
	 \includegraphics[width=.4\textwidth, trim={0.25cm 0.2cm 0.25cm 0.25cm}, clip]{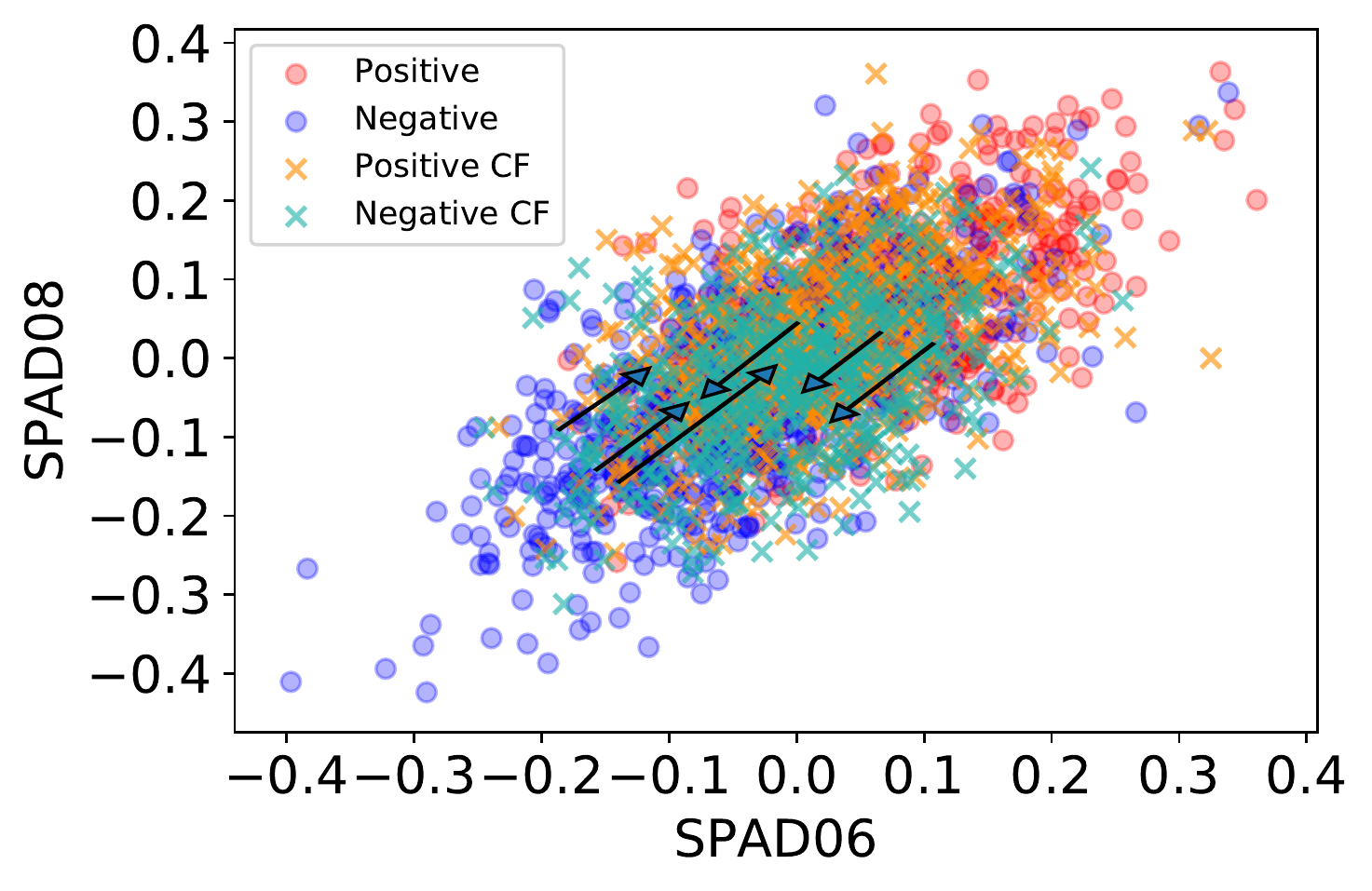}\\
	 \specialrule{0em}{-5pt}{-1pt}
	 \footnotesize (c) Ours
	\end{tabular} 
	
	\caption{Visualization of related variables in Sangiovese. Positive and negative indicate the predictions.}
	\label{sangiovese_vis}
\end{figure}

\noindent\textbf{Simulated Bayesian Network Dataset.} 
The dataset includes 10000 samples with 14 features and a categorical output for quality. We follow the data splitting setting in \cite{mahajan2019preserving}. We measure two monotonic relationships \textit{(1) between SproutN and BunchN where if SproutN increases/decreases, BunchN increases/decreases; (2) between SPAD06 and SPAD08, where the same trend hold}.

The quantitative results are shown in Table \ref{tab:performance2}. As previously, the models achieve around near 100\% validity. Our model is the second best in terms of Euclidean distance. However, in terms of Mahalanobis distance, our method has outperformed other methods. Besides, the constraint feasibility score of our method is nearly 100\%, indicating the model perfectly captures the correlations between the variables. Figure \ref{sangiovese_vis} shows that the counterfactuals generated by EB-VAE of both labels tend to cluster together while the proposed model still keeps a clear boundary between the sets. Most of the changes that samples undergo in order to flip the labels are along the direction representing the relationship between the attributes.

\subsection{Results on Diabetes Dataset (RQ2)}

\noindent\textbf{Dataset description.} The dataset collected by \cite{smith1988using} contains medical records for Pima Indians with and without diabetes. Overall, the dataset contains 768 records, out of which 268 are diabetic and 500 are not. The records that contained missing values were removed, resulting in 336 records.  Each record consists of 8 attributes. We selected the following 7: number of times pregnant, plasma glucose concentration at 2 hours in an oral glucose tolerance test, diastolic blood pressure, triceps skin fold thickness, 2-hour serum insulin, body mass index (BMI), and age. The target variable is the patients' diabetic conditions i.e. positive (diabetic) or negative (not).

\begin{figure}
    \includegraphics[width=1\textwidth, trim={0.0cm 5.0cm 0.0cm 5.0cm},clip]{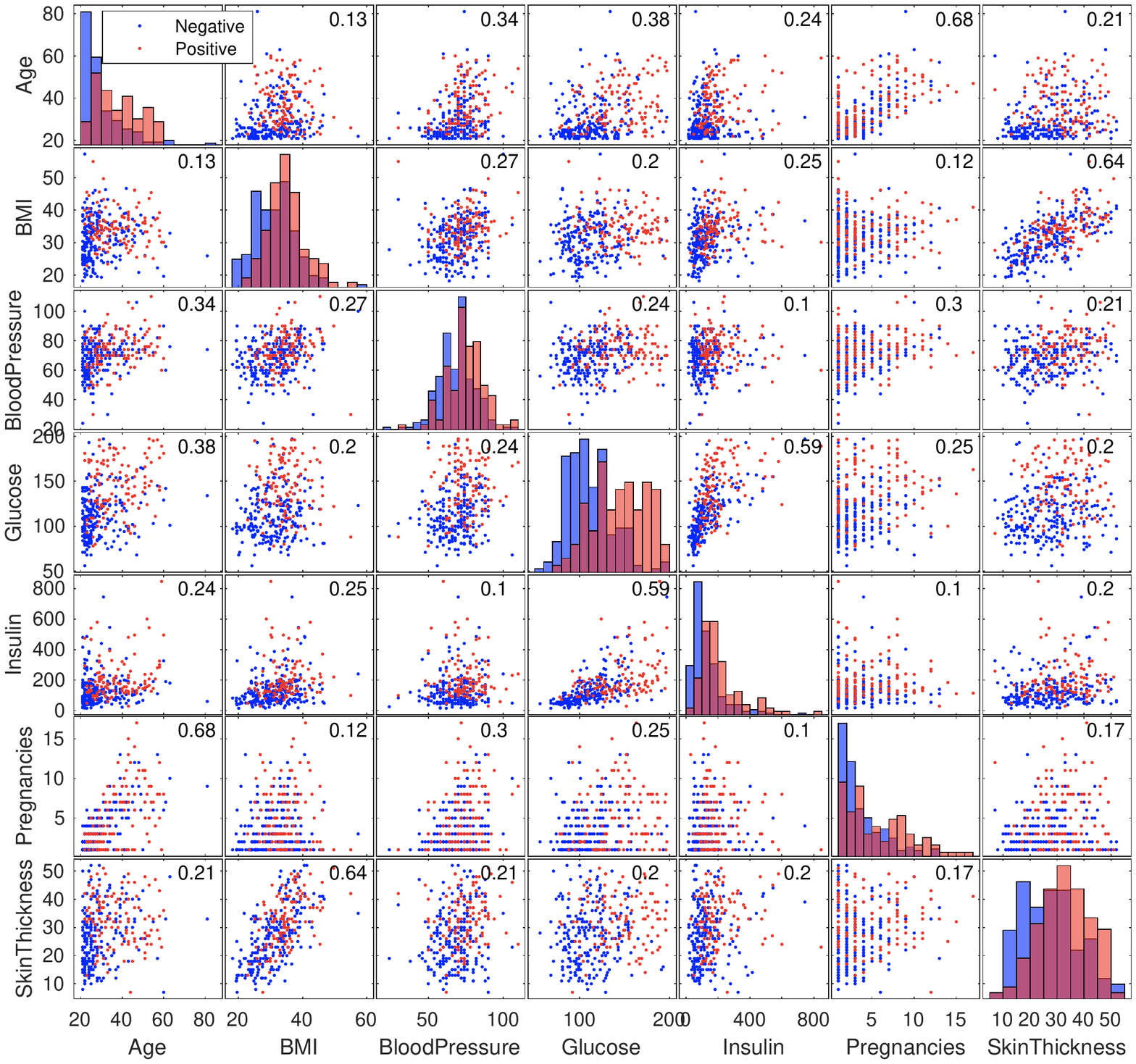}
    	\caption{Visualization of paired attributes with correlations shown in top-right corners of corresponding panes. The positive class (red) indicates the diabetic patients while the negative (blue) is healthy.}\label{data_diabete}
\end{figure}

Before proceeding to generating and testing counterfactuals, we first conducted a qualitative analysis of the dataset in order to compare the observed relationships to prior domain knowledge on diabetes. Diabetes is a disease that is caused by either pancreas not being able to produce enough insulin or by organism not being able to use the produced insulin.
Figure \ref{data_diabete} shows pairwised correlations, it is easy to see that most pairs exhibit positive correlation. In particular, there is a strong linear relationship between insulin and glucose and hence in order to change the label from positive to negative, both glucose and insulin should be decreased.
Further, it is easy to observe that besides having thinner skin, young people are less likely to be diabetic, and hence the counterfactuals that flip patients with diabetes to healthy might require to decrease their age. Given that decreasing age is infeasible, the changes that decrease the age are penalised. This prior expert knowledge is important in order to generate meaningful counterfactuals. It is also known that blood pressure and BMI are positively correlated, which is also observed in the chart ($\rho = 0.27$).  Interestingly, the number of pregnancies is correlated with blood pressure($\rho = 0.3$),  moreover women with a larger number of pregnancies are slightly more likely to have diabetes, which implies that the counterfactual might result in a decreased number of pregnancies in order to flip the predicted label from diabetic to healthy, such changes should also be penalised. Another interesting aspect that follows from the analysis of figure \ref{data_diabete} is that blood pressure does not well separate healthy cohort and patients with diabetes, and hence should not undergo a significant change in the generated counterfactual compared to the original sample.

	
	
	

We measure the constraint feasibility score by the pairs of blood pressure and BMI since from the domain knowledge we know that blood pressure and BMI are positively correlated ($\rho=0.27$), and hence in a counterfactual, the blood pressure and BMI should both increase or decrease. Since the number of samples is small, we use leave-one-out algorithm, which takes one sample for testing the model, while the rest of samples are used in training, this procedure is repeated over all samples, and an average accuracy is computed. 

\begin{figure}
	\centering
	\begin{tabular}{@{}c@{}c@{}}
		\includegraphics[width=.4\textwidth, trim={0.0cm 0.25cm 0.25cm 0.25cm}, clip]{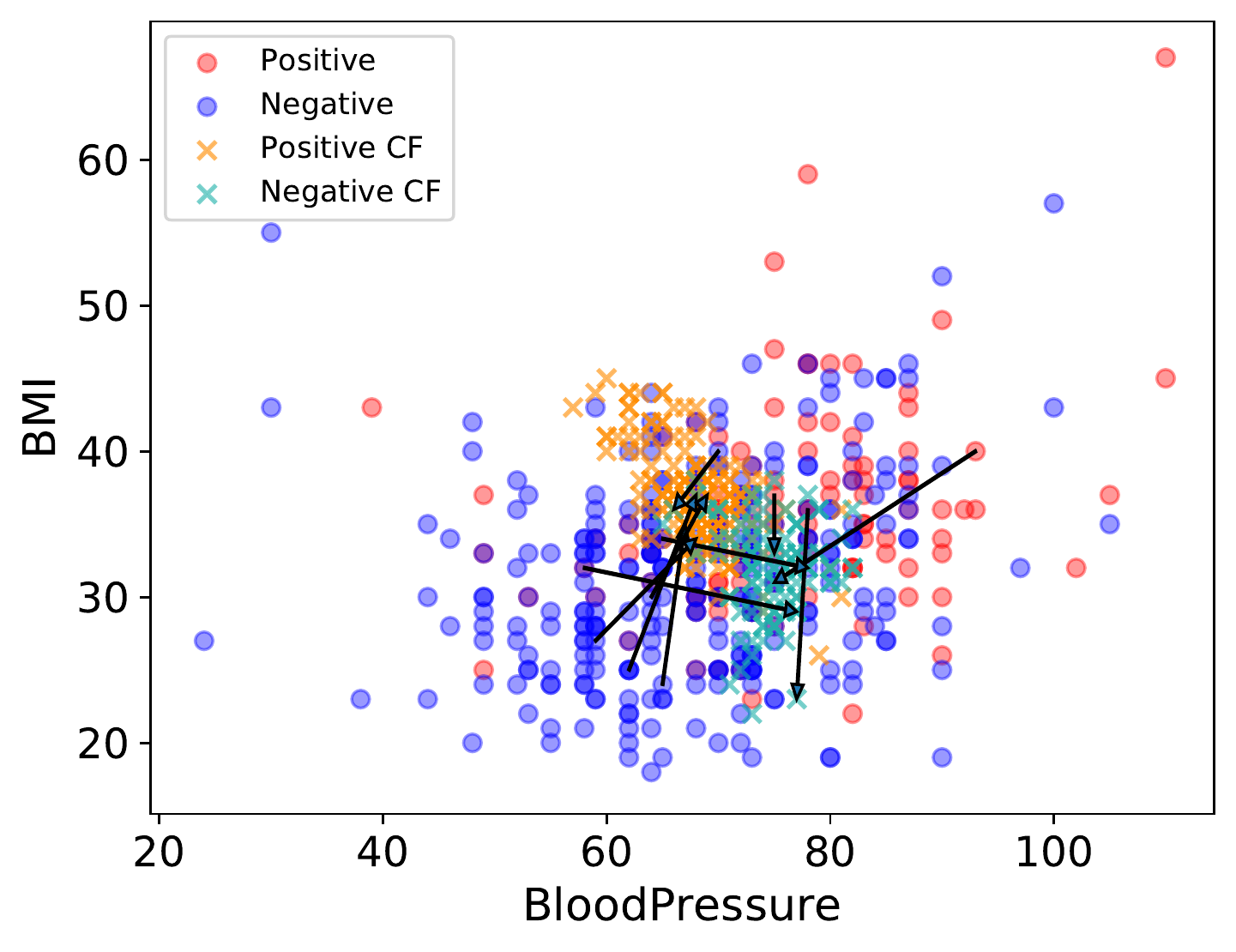} & \includegraphics[width=.4\textwidth, trim={0.0cm 0.25cm 0.25cm 0.25cm}, clip]{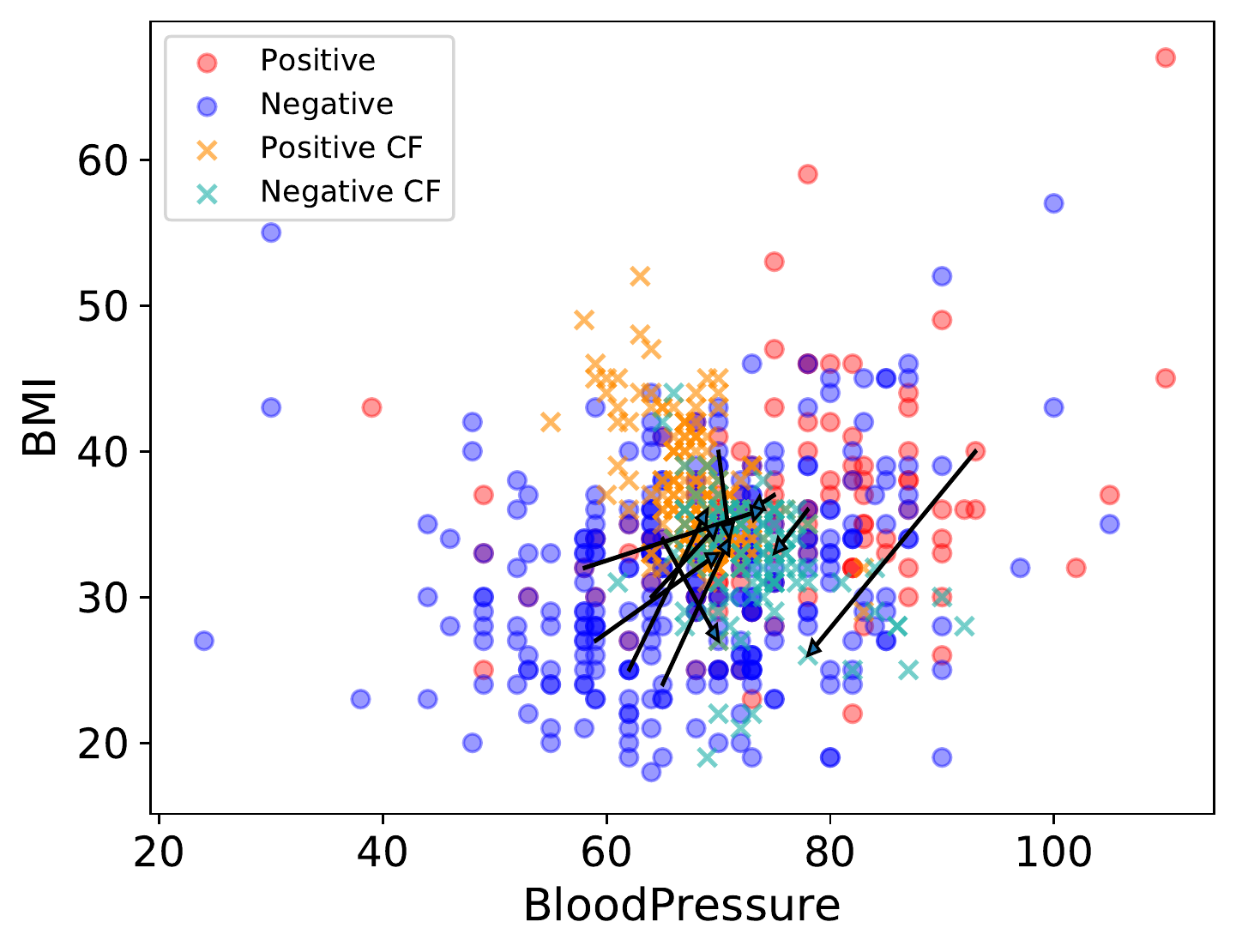} \\
		\footnotesize (a) CF-VAE &\footnotesize (b) EB-VAE
	\end{tabular}
	
	\begin{tabular}{@{}c@{}c@{}}
		\includegraphics[width=.4\textwidth, trim={0.0cm 0.25cm 0.25cm 0.25cm}, clip]{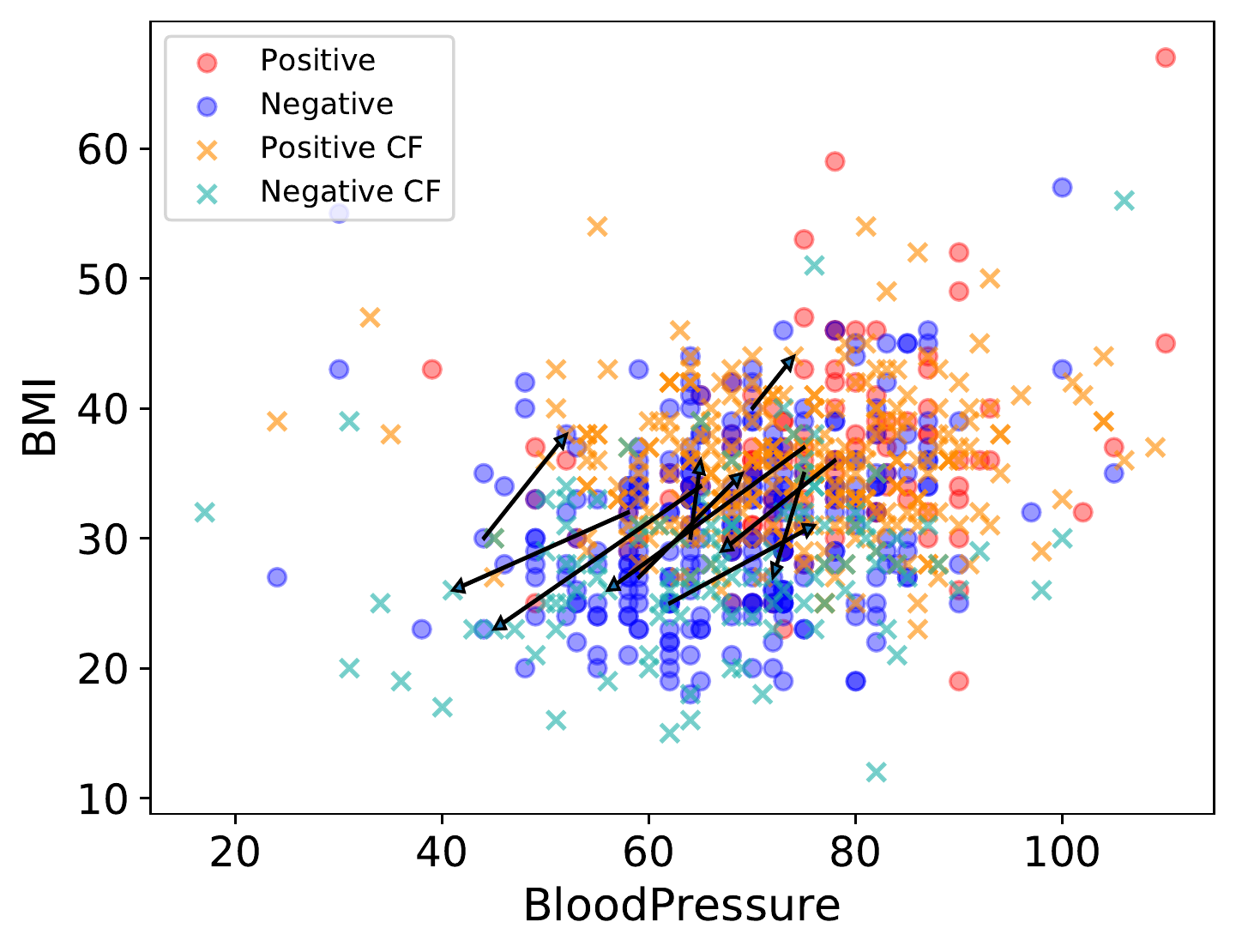} \\ \vspace{-2pt}
		\footnotesize (c) Ours 
	\end{tabular}

	\caption{Visualization of baseline and our method on diabetes dataset. Positive CF indicates the generated counterfactuals which are predicted to have diabetes while negative CF indicates counterfactuals which are predicted to be healthy.}
	\label{diabete_vis_bmi}
\end{figure}

The quantitative evaluations are presented in Table \ref{diabete}. The validity for the methods are all nearly 100\% and is not informative so we did not put them in the table. As it can be seen, around 74\% of the counterfactuals satisfy the positive relationship between blood pressure and BMI. While the best baseline method attains about 60\% counterfactuals that satisfy the relationship. The Mahalanobis distance of our method are significantly smaller than those of all the other baseline models. 
To qualitatively compare the methods, we plotted the counterfactuals on the blood pressure vs BMI (fig. \ref{diabete_vis_bmi}). Given that blood pressure and BMI are positively correlated, we expected the counterfactuals to maintain this linear relationship, and hence the changes in these attributes should be along the direction representing the correlation. While our model maintained the correlation, the generated counterfactuals for other baselines, did not preserve the correlation, qualitatively demonstrating the efficiency of our method in learning the internal relationships between the two attributes and generating proper counterfactuals. It should also be noted that the class label could flip due to the change in other patient's characteristic while keeping blood pressure and BMI constant. 

\begin{table}[t]
	\centering
	\captionof{table}{Results on Diabetes dataset.}
	\scalebox{1}{\begin{tabular}{lcccc}
		\toprule
		Methods & Valid (\%) & Cont (\%) & Euclidean Dist & Mahalanobis Dist \cr
		\midrule
		CF-VAE & 97.70 & 58.16 & 0.1288 & 3.2385\cr
		EB-VAE & 99.74 & 60.97 & 0.1337 & 3.3189\cr
		Ours   & 99.74 & 88.44 & 0.1279 & 2.4042\cr
		\bottomrule
	\end{tabular}}
	\label{diabete}
\end{table}

\begin{figure}[t]
	\centering
	\begin{tabular}{@{}c@{}c@{}}
	  \includegraphics[width=.4\textwidth, trim={0.0cm 0.30cm 0.0cm 0.3cm}, clip]{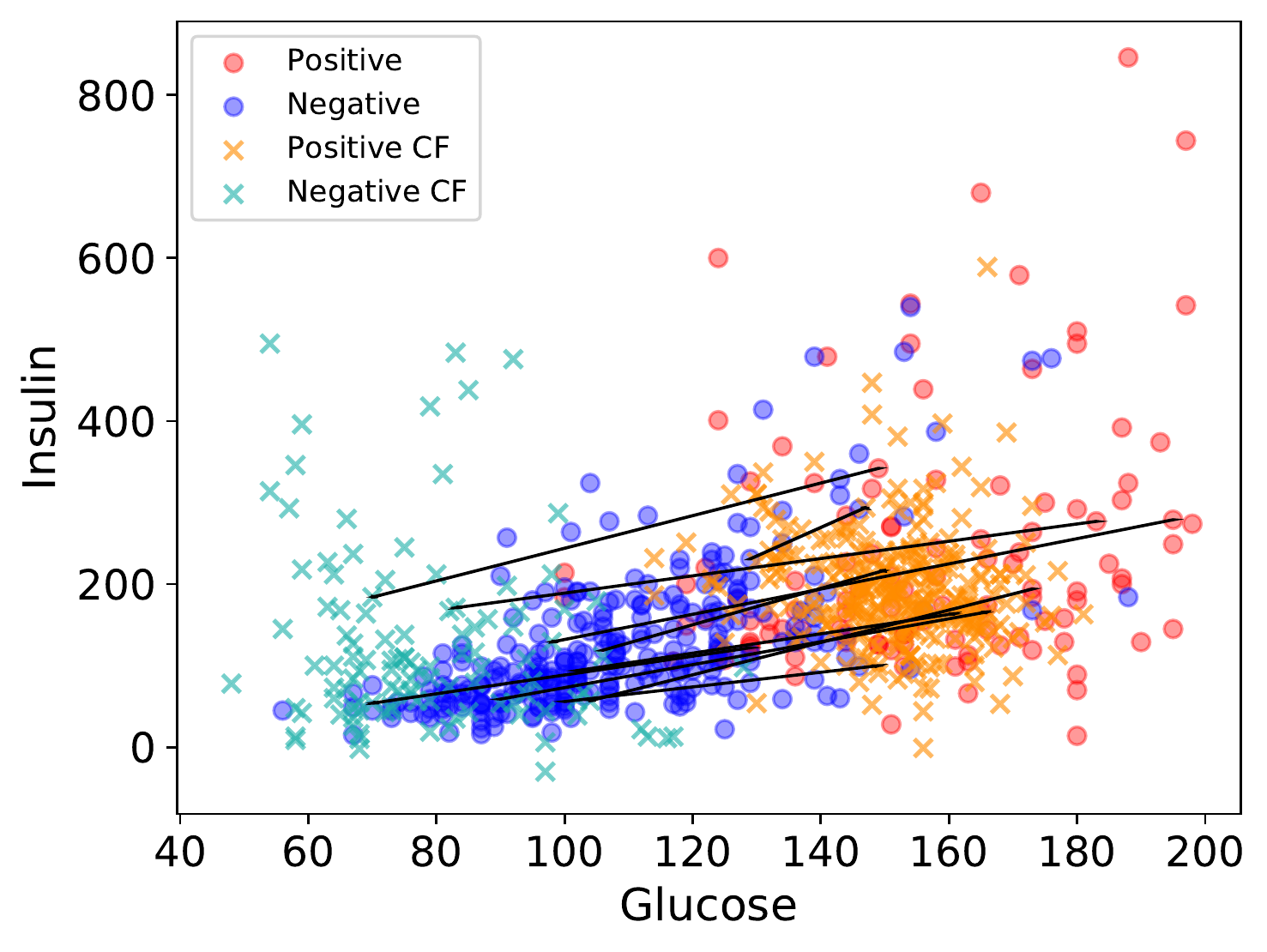} 
	  \includegraphics[width=.4\textwidth, trim={0.0cm 0.30cm 0.0cm 0.3cm}, clip]{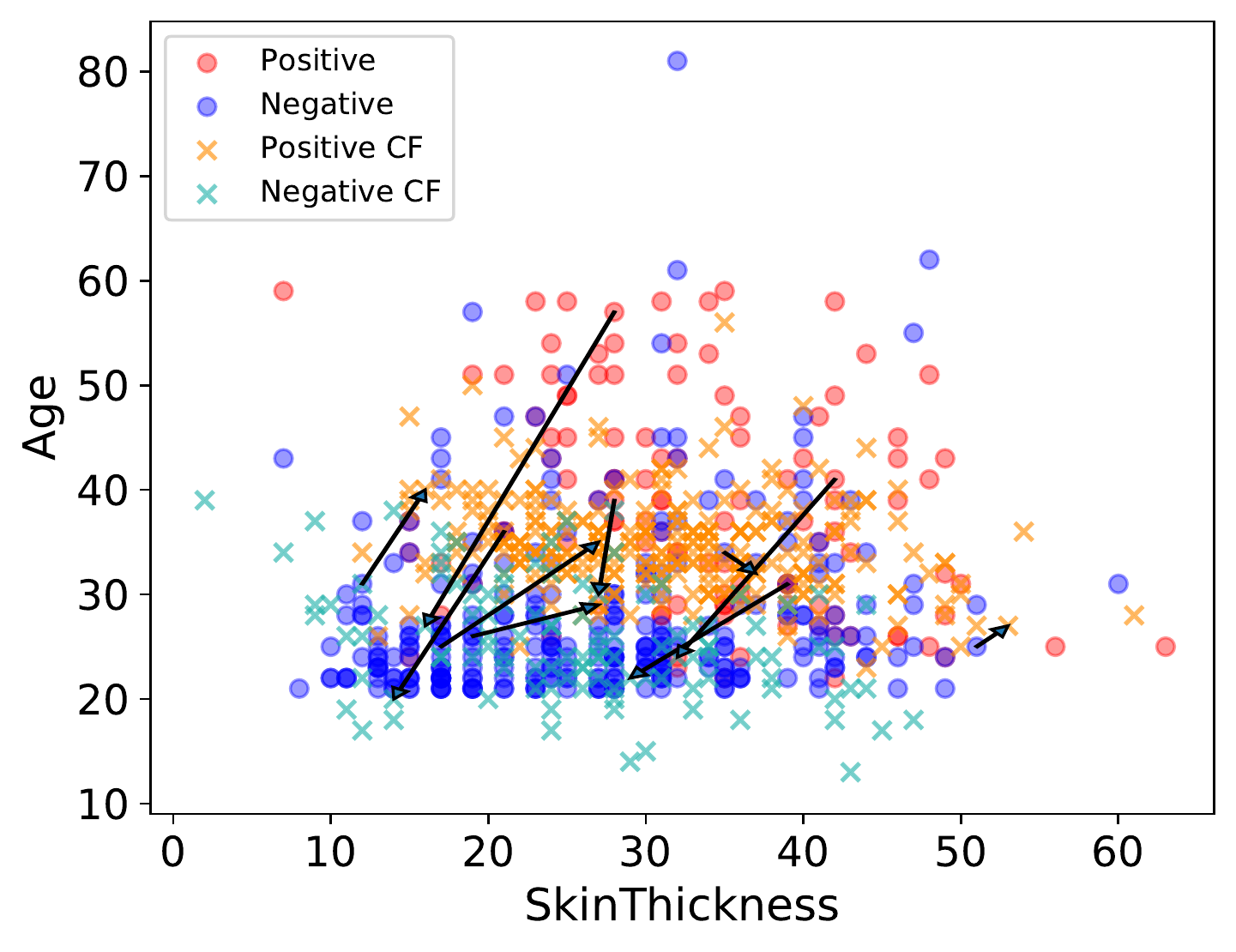} \\
	\end{tabular}
	
	
	\caption{Visualization of some paired attributes in our method.}\label{data_diabete_other}
\end{figure}

	
	


We further provide two examples of generated counterfactuals (table \ref{tab:diabete_examples}) in order to demonstrate a use-case of our model being useful in consulting patients. In both examples, our model suggests to decrease glucose and insulin, following our domain knowledge that glucose directly contributes to diabetes. Besides, it can be observed that the blood pressure, and skin thickness do decrease, which matches our observations in figure \ref{data_diabete}, as again, blood pressure and skin thickness seem to be associated with diabetes and correlated with each other ($\rho=0.21$). Similarly, the decrease in both blood pressure and BMI, demonstrates that the model learns the correlation between the two attributes as well as their associations with diabetes. This insight could be used by medical experts to provide medical advice to the patients. Example 2 illustrates a limitation of our method. The model suggests reducing age, even though a constraint was added to penalise unacceptable changes. In particular, nearly a third of the counterfactuals are generated with a slightly decreased age. This is especially true for counterfactuals flipping diabetes to healthy. We see that in generating counterfactuals for healthy, about 88\% of the counterfactuals chose to increase the age. However, for diabetic counterfactuals, almost all counterfactuals resulted in a decreased age. The model often generates such counterfactuals due to a strong correlation between age and the target value i.e. young people are less likely to have diabetes (fig. \ref{data_diabete}). Since the objective of this paper is to study the relationships between attributes, we will leave the unary (attribute that can only change in one direction) constraint for future work. 

To summarize, our method successfully captures the relationship between attributes (specifically the blood pressure and BMI). The generated counterfactuals preserve the relations that exist in the data and can be a good reference to suggest a patient what needs to be changed to become healthy. While we found age and pregnancies sometimes decrease, we will leave it to future work and constrain the changes so that such attributes should either stay constant or could only change in one direction.

\begin{table*}[t]
	\centering
	\caption{Examples of explanations with our method on Diabete dataset.}
	\begin{tabular}{lcccccccc}
	\toprule
	Attributes & Pregnancies &Glucose & Blood Pressure & Skinthickness & Insulin & BMI & Age & Diabetic
	\cr
	\midrule
	\midrule
	Original Input  & 1 & 166  & 73   & 17   & 144& 23 & 33 & Yes \cr
	Counterfactual  & 1 & 113 & 54 & 7 & 13 & 18 & 34 & No  \cr
	\midrule
	Original Input  & 0 & 134  & 93   & 46 & 145 & 40 & 26 & Yes  \cr
	Counterfactual  & 0 & 73 & 89 & 42 & 106  & 38 & 18 & No  \cr

	\bottomrule
	\end{tabular}
	\label{tab:diabete_examples}
\end{table*}

\section{Conclusion and Future Work}

In this paper, we proposed a novel method for generating counterfactuals that preserves the relationships between variables learned from data. A VAE was trained to reconstruct input vectors. Then a modulation network was trained to perturb the latent space representations such that counterfactuals are generated in place of input vectors. We experimented with synthetic and real datasets. The results show the superiority of the proposed method in generating realistic counterfactual explanations that preserve relationships while explaining machine learning models.

The proposed approach can be further extended as follows: (1) While we considered only binary classification tasks, it is of practical importance to extend the proposed model in a more complex multiple-label settings. (2) The model could further be modified to generate several counterfactual from a single sample by adding Gaussian noise to the latent space. (3) From a practical perspective, we plan to apply the proposed model to datasets of electronic medical records of patients with various diseases. The proposed model will provide medical experts with counterfactual explanations that will help patients to make necessary but minimal changes in either their lifestyles or treatments to further improve a probability of a positive health outcome. 

\section*{Acknowledgements}
Artem Lenskiy was funded by Our Health in Our Hands (OHIOH), a strategic initiative of the Australian National University, which aims to transform healthcare by developing new personalised health technologies and solutions in collaboration with patients, clinicians, and health care providers.



\bibliographystyle{unsrtnat}
\bibliography{references}






\end{document}